\documentclass[runningheads]{llncs}

\usepackage{eccv}

\usepackage{eccvabbrv}

\usepackage{graphicx}
\usepackage{booktabs}
\usepackage{arydshln}
\usepackage[accsupp]{axessibility}  % Improves PDF readability for those with disabilities.

\usepackage{xcolor}
\usepackage{multirow, makecell}
\definecolor{tabhighlight}{HTML}{e5e5e5}
\usepackage{color, colortbl}
\usepackage{tikz}
\usepackage{pifont}
\usepackage{fontawesome}
\usepackage{wrapfig,lipsum,booktabs}

\usepackage[misc]{ifsym}
\newcommand{\ours}{\color{black}Plain-Det}
\definecolor{scratch}{HTML}{001219}
\definecolor{pretrain}{HTML}{0A9396}
\definecolor{obj}{RGB}{255,143,102}
\definecolor{coco}{RGB}{174,42,59}
\newcommand{\scratch}{\textcolor{scratch}{$\mathbf{\circ}$\,}}
\newcommand{\pretrain}{\textcolor{pretrain}{$\bullet$\,}}
\newcommand{\cmark}{\ding{51}}%
\newcommand{\xmark}{\ding{55}}%
\newcommand{\tablestyle}[2]{\setlength{\tabcolsep}{#1}\renewcommand{\arraystretch}{#2}\centering\footnotesize}
\DeclareUnicodeCharacter{2206}{\increment}
\definecolor{green}{HTML}{1e8e3e}
\usepackage{hyperref}

\usepackage{orcidlink}
\newcommand\blfootnote[1]{%
  \begingroup
  \renewcommand\thefootnote{}\footnote{#1}%
  \addtocounter{footnote}{-1}%
  \endgroup
}
\begin{document}

% \title{Plain-D$^{\text{N}}$et: Three Things Everyone should know \\ about Multi-Dataset Detection} 
\title{Plain-Det: A Plain Multi-Dataset \\ Object Detector} 
\titlerunning{Plain-Det}

\author{Cheng Shi$^*$ \and
Yuchen Zhu$^*$ \and
Sibei Yang\textsuperscript{\Letter}}

\authorrunning{Cheng et al.}

\institute{School of Information Science and Technology\\
ShanghaiTech University\\
\email{\{shicheng2022,yangsb\}@shanghaitech.edu.cn}\blfootnote{$*$ Equal contribution. \Letter~Corresponding author.}\\ 
 }

\maketitle
\begin{abstract}
Recent advancements in large-scale foundational models have sparked widespread interest in training highly proficient large vision models. A common consensus revolves around the necessity of aggregating extensive, high-quality annotated data. 
However, given the inherent challenges in annotating dense tasks in computer vision, such as object detection and segmentation, a practical strategy is to combine and leverage all available data for training purposes.
In this work, we propose Plain-Det, which offers flexibility to accommodate new datasets, robustness in performance across diverse datasets, training efficiency, and compatibility with various detection architectures. 
We utilize Def-DETR, with the assistance of Plain-Det, to achieve a mAP of 51.9 on COCO, matching the current state-of-the-art detectors.
We conduct extensive experiments on 13 downstream datasets and Plain-Det demonstrates strong generalization capability.
Code is release at \url{https://github.com/ChengShiest/Plain-Det}.
\keywords{Object detection \and Multiple datasets \and Proposal generation }

\end{abstract}    
\section{Introduction}
\label{sec:intro}

Large-scale datasets have fostered significant advances in computer vision, ranging from ImageNet~\cite{deng2009imagenet} for image classification to more recent datasets~\cite{coco,gupta2019lvis} like SA-1B~\cite{sam} for image segmentation. 
Object detection~\cite{ren2015faster,redmon2017yolo9000,detr}, as one of the fundamental tasks in computer vision, inherently demands large-scale annotated data. However, annotating such extensive and densely annotated objects is both costly and challenging. % but obtaining such large-scale dense annotated data is both costly and challenging.
%However, annotating dense objects in images is not only costly but also challenging to define a unified and standard annotation granularity of semantics. 
Another straightforward and practical approach is unifying multiple existing object detection datasets~\cite{coco, gupta2019lvis,shao2019objects365} to train a unified object detector~\cite{unidet, scaledet, detectionhub}.  
%Nonetheless, inconsistent taxonomies and training strategies between datasets pose new challenges: as more datasets are unified, the performance of multi-dataset object detection across multiple datasets does not consistently improve and sometimes even falls below the results achieved with a single dataset. 
Nonetheless, inconsistency between datasets, such as differing taxonomies and data distributions as illustrated in Fig~\ref{fig:fig1}\textcolor{red}{(a)}, poses challenges to multi-dataset training.  %\textcolor{red}{perf not satisfied?}

%In this paper, we aim to probe the pivotal factors %influencing the success of multi-dataset object detection 
%and offer insights 
%In this paper, we attempt to address the question: how can we 
In this paper, we aim to address the challenges to train an effective and unified detector using multiple object detection datasets, with the expectation that it should: (1) \textbf{\textit{Flexibility to new datasets}} in a seamless and scalable manner without requiring manual adjustments, complex designs, and training from scratch. (2) \textbf{\textit{Robustness in performance}} when incrementally incorporating new datasets, consistently leading to improved performance or, at the very least, maintaining stable performance. (3) \textbf{\textit{Training efficiency}}. The number of training iterations required for multi-dataset training is no greater than that for a single dataset. (4) \textbf{\textit{Compatibility with detection families}}, such as Faster-RCNN series~\cite{ren2015faster,girshick2015fast,sparsercnn} and DETR-based detection architecture~\cite{zhu2020deformable,detr,dino}.

% \begin{figure}[t]
%     \centering
%     \caption{XXXX.}
% \includegraphics[width=0.95\textwidth]{fig/fig1_v3.pdf} 
%     \vspace{-2mm}
% \label{fig:fig2} 
% \end{figure}

\newcommand{\fig}[2][1]{\includegraphics[width=#1\linewidth]{#2}}
\newcommand{\figh}[2][1]{\includegraphics[height=#1\linewidth]{fig/#2}}
\newcommand{\figa}[2][1]{\includegraphics[width=#1]{fig/#2}}
\newcommand{\figah}[2][1]{\includegraphics[height=#1]{fig/#2}}

%------------------------------------------------------------------------------
% \vspace{-6pt}
\begin{figure*}[t]
\captionof{figure}{\textbf{The benefits and challenges of multi-dataset object detection.}
(a) Various datasets span diverse taxonomies and data distributions. (b) Semantic space calibration. (c) Our approach leverages the advantages of training across multiple datasets to achieve performance enhancements through scaling up data volume.
}
\vspace{-2mm}
% \scriptsize
% \begin{tabular}{ccc}
% \fig[.4]{fig/fig1V5_p1} &
% \fig[.32]{fig/fig1V5_p2.pdf}&
% \fig[.2405]{fig/fig1V5_p3.pdf}
\fig[.99]{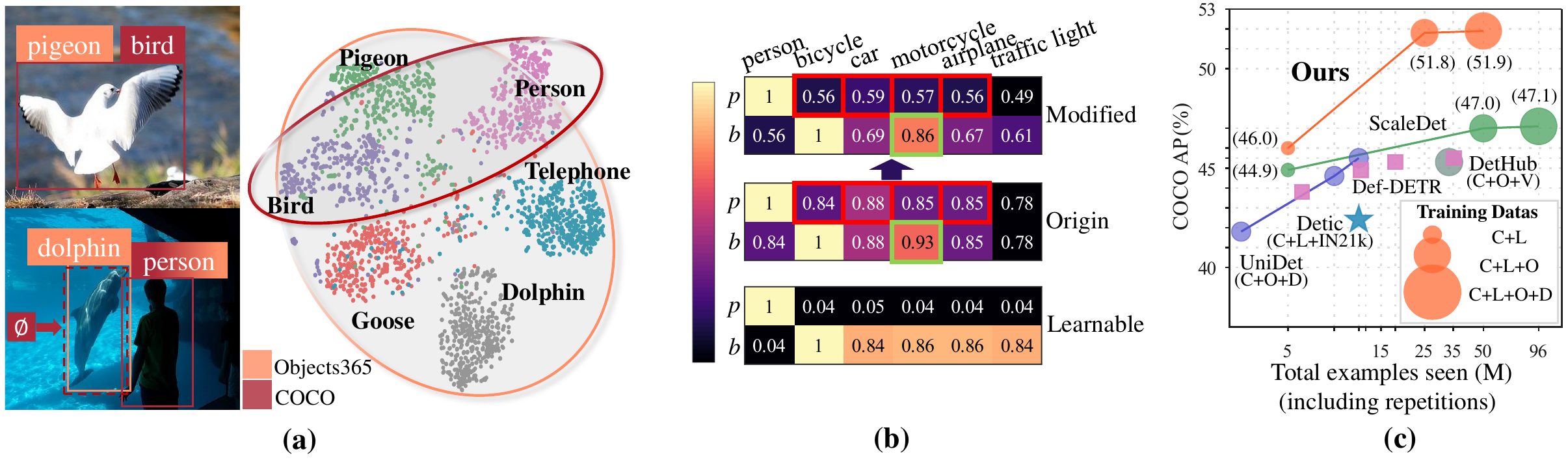}\\

% (a) & (b) & (c) \\
% \makecell{(a) Sampled Images and Features\\ in  \textcolor{coco}{COCO} \& \textcolor{obj}{Obj365}} &
% \makecell{(c) The performance-data volume \\ trade-off curve.}\\

% \end{tabular}%
\label{fig:fig1}
\vspace{-5mm}
% \par\vspace{24pt}
\end{figure*}

To start, we introduce a simple yet flexible multi-dataset object detection baseline, which boldly challenges some recent design principles while keeping other advances. 
Recent works~\cite{detectionhub, scaledet, unidet} explicitly unify taxonomies across different datasets into a single, unified one. However, despite their automatic methods, they still require carefully hand-designed components and lack flexibility in scaling to more datasets. This is primarily because 1) the mapping from dataset-specific label spaces to a unified one, learned automatically, becomes noisier as the label space size grows, and 2) incorporating new datasets necessitates reconstructing the unified taxonomy. %\TODO{Figure xx shows label mapping xxxx}
Therefore, \textit{\textbf{we introduce a shared detector with entirely dataset-specific classification heads to naturally prevent conflicts between different taxonomies and ensure flexibility.}} 
Furthermore, following~\cite{vild, zhou2022detecting, scaledet}, we utilize text embeddings of category labels to build a shared semantic space of all labels. Notably, the semantic space implicitly establishes connections between labels from different classifiers, enabling full use of all the training data despite the dataset-specific classification head. 
%Third, we combine the class-aware repeat factor sampling strategy~\TODO{cite} and multi-dataset sampling strategy~\TODO{cite} to better balance class distributions and dataset sizes, following \TODO{cite unidet}. 
%without any extra design of unifying label space and dataset-aware 
%\textcolor{red}{
%While our baseline model is flexible, its performance in multi-dataset object detection is inferior to the model employing shared class classification with a unified label space, as indicated by a $-0.5\%$ mAP when comparing line 1 and line 2 in Table~\ref{table:ablation}.}
Although our multi-dataset baseline model demonstrates flexibility, its performance notably lags behind that of the single-dataset object detector, with a $-3.2\%$ mAP decrease observed when comparing our baseline ``\scratch C+L+O+D'' and ``\scratch single'' model in Table~\ref{table:increasing_multi}.

%------------------------------------------------------------------------------
% \vspace{-6pt}
\begin{figure*}[t]
\captionof{figure}{\textbf{The insights for sparse proposal generation and emergent property.} (a) Difference between dense proposal generation and sparse proposal generation.
(b) Analysis of two types of proposal generation under multi-dataset object detection training. (c) The emergent property in multi-dataset training. The detector trained on COCO+O365+LVIS shows unstable performance on LVIS.
}
\vspace{-2mm}
% \scriptsize
\begin{tabular}{ccc}
\fig[.24]{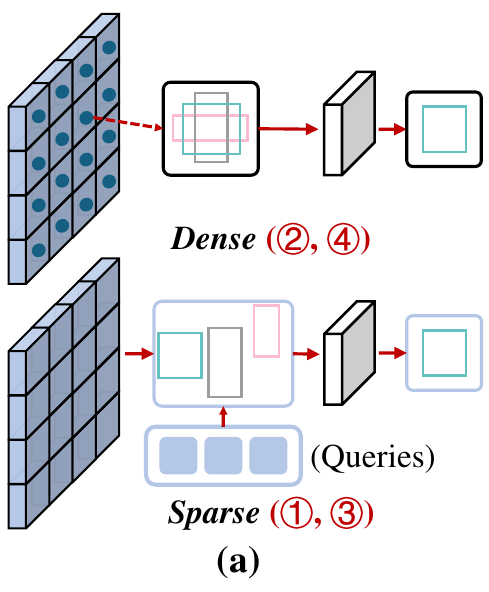} &
\fig[.34]{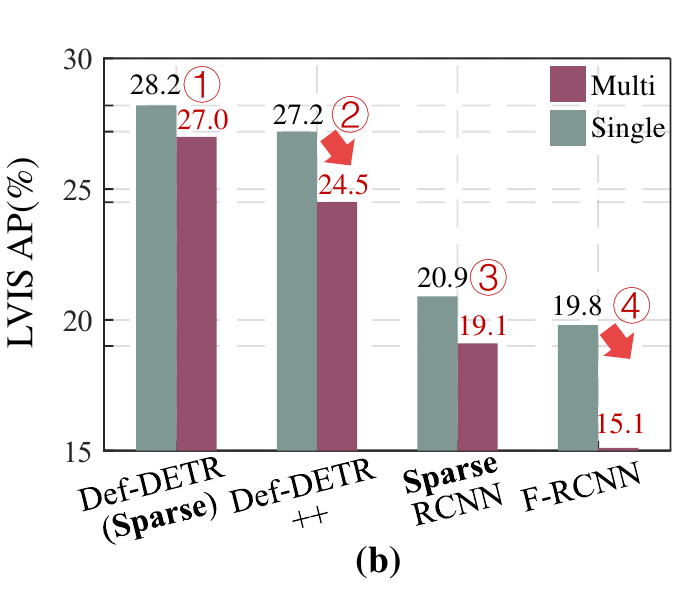} &
\fig[.39]{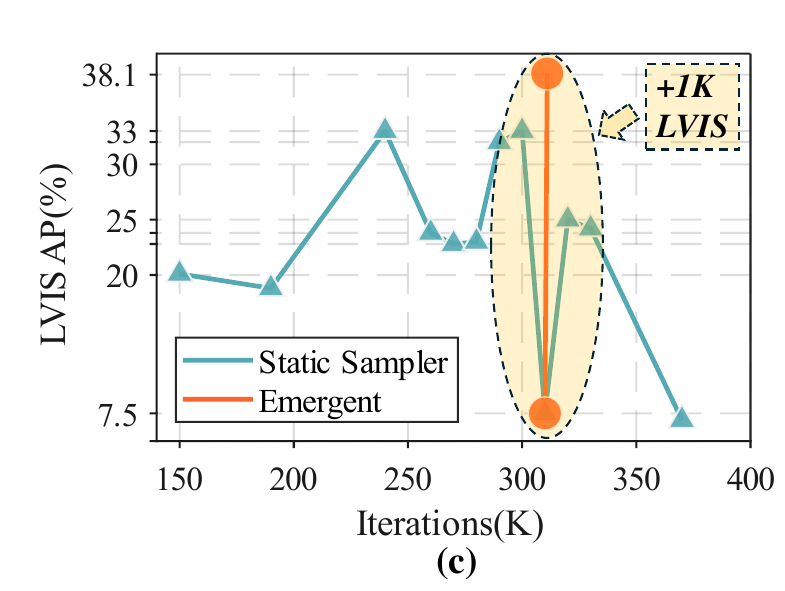} \\

% (a)& (b)  & (c)\\
\end{tabular}%

\label{fig:fig2}
% \par\vspace{24pt}
\end{figure*}

Therefore, we probe the pivotal factors impacting the success of the baseline and offer three insights to empower it to be not only super flexible but also highly effective:

    \noindent\textbf{\textit{1) Semantic space calibration}} %Text embeddings are used as classification weights for building the common semantic space for different dataset-specific classifiers. However, is the classifier with frozen text embeddings perfect for object detection? 
    %Although the text embeddings have been used as classification weights, 
    is inspired by questioning whether the classifier with frozen text embeddings is perfect for object detection. 
    \textcolor{black}{Fig~\ref{fig:fig1}\textcolor{red}{(b)}-``origin''} shows the similarity matrix of text embeddings between categories, which is noticeably different from the one generated by learnable classification weights (Fig~\ref{fig:fig1}\textcolor{red}{b}-``learnable'').  
    The bias originates from CLIP’s training data distribution; for instance, the text-image pairs in CLIP typically exhibit a long-tail distribution in the frequency of nouns. This results in a high similarity between the text embeddings of frequently occurring nouns (such as `person' in Fig 1b) and other words (including \texttt{NULL}). In turn, we discover that an infrequently occurring \texttt{NULL} exhibits high similarity with frequently occurring words and low similarity with infrequently occurring ones. Therefore, we can treat the empty string \texttt{NULL} as a meaningless basis to extract the basis driven by frequency, resulting in the calibrated similarity matrix shown in Fig~\ref{fig:fig1}\textcolor{red}{(b)}-``modified''.
 
    \noindent\textbf{\textit{2) Sparse proposal generation}}. 
In object detection, object proposal generation is crucial, especially in multi-dataset scenarios. This is because the same object proposals \textcolor{black}{used as anchors to} predict different object sets for different datasets. For example, while COCO and LVIS share the same image set, there are significant differences in annotated categories. This necessitates that the same object proposals within the same image can \textcolor{black}{anchor} different objects from both COCO's 80 categories and LVIS's 1203 categories. Currently, object proposal generation methods can be broadly categorized into two types~\cite{sparsercnn}: 1) dense or dense-to-sparse proposal generation~\cite{redmon2017yolo9000,ren2015faster}, which generates proposals across all image grids or selects a small subset from dense proposals, and 2) sparse proposal generation~\cite{detr,sparsercnn,zhu2020deformable}, which typically directly generates a set of learnable proposals (see Fig~\ref{fig:fig2}\textcolor{red}{a}). Therefore, we conduct preliminary experiments and comparisons of these two types of proposal generation methods in multi-dataset object detection on the COCO and LVIS datasets. The results suggest that sparse proposal generation methods consistently outperform the dense approach across both object detector families, as shown in Fig~\ref{fig:fig2}\textcolor{red}{b}. 
%\textcolor{blue}{One possible reason is that the image feature map only contains information about objects (such as shape, size, and category), making it difficult to capture the dataset distribution. During training on a single dataset, since annotations are consistent, there is no need to explicitly model dataset-specific information. However, during training on multiple datasets, although the same image generates the same feature map, different datasets require different sets of proposals. This necessitates the additional modeling of dataset information within the parameters of the backbone network, thereby increasing training complexity. In contrast, sparse queries can capture the distribution of the dataset, making it easier to learn from multiple datasets. However, the performance of multi-dataset training still falls below that of single-dataset training, due to the need for the same query to capture the priors of different datasets. Therefore, we improve sparse query to class-aware query based on the unified semantic space and image prior, which mitigates the challenge of a set of queries having to accommodate multiple datasets.}
One possible reason is that compared to dense proposal generation, sparse proposals (\ie, sparse queries)~\cite{detr,dndetr,dino,zhu2020deformable} have been demonstrated to capture the distribution of the dataset, making it easier to learn the joint distribution from multiple datasets. However, the performance of multi-dataset training still falls below that of single-dataset training, due to the need for the same queries to capture the priors of different datasets. Therefore, we improve sparse queries to class-aware queries based on the unified semantic space and image prior, which mitigates the challenge of a set of queries having to accommodate multiple datasets.

    \noindent\textbf{\textit{3) Dynamic sampling strategy inspired by the emergent property}}. Despite the two insights above unlocking the potential of training a unified detector on multiple datasets like COCO~\cite{coco} and LVIS~\cite{gupta2019lvis}, the inclusion of dataset Objects365~\cite{shao2019objects365} leads to large fluctuations in the detection performance during training (see Fig~\ref{fig:fig2}\textcolor{red}{b}-``static sampler''), primarily due to noticeable imbalances in dataset sizes (see Fig~\ref{fig:fig2}\textcolor{red}{c}). Surprisingly, we observe that even when the detector in a given iteration has low precision on a dataset, it can substantially enhance its precision by undergoing a few additional training iterations on that specific dataset (see Fig~\ref{fig:fig2}\textcolor{red}{b}-``emergent''). We attribute this phenomenon to an emergent property of multi-dataset detection training: a detector trained on multiple datasets inherently possesses a more general detection capability than training on a single dataset, and the ability can be activated and adapted to the particular dataset by a few dataset-specific iterations. Inspired by the property, we propose a dynamic sampling strategy to achieve better balance among different datasets, which dynamically adapts the multi-dataset sampling strategy in subsequent iterations based on the dataset-specific loss observed previously. 

%\end{itemize}
%(1) \textbf{\textit{Sparse xxxxx.}}  (2) \textbf{\textit{Dynamic sampling strategy inspired by the emergent property}}. (3) \textbf{\textit{Semantic space regularization.}} 

%
% \input{tables/table1}

Finally, we introduce Plain-Det, a simple yet effective multi-dataset object detector that can be easily implemented by directly applying the three proposed insights to the baseline, thanks to the baseline's flexibility. %\TODO{Thanks to the flexibility of our model, the insights are easy xxx on it}
%兼具灵活性，让各分支处理避免冲突； 鲁棒性：dynamic sampling策略； 有效性：充分利用数据，通过queruies来xxxx且通过更有判别能力的text embedding来xxx；
In summary, our contributions are: 
\begin{itemize}
\setlength{\itemsep}{0pt}
\setlength{\parsep}{0pt}
\setlength{\parskip}{0pt}
%\item We offer three key insights to unlock the challenges of multi-dataset object detection training, including the calibration of the label space, the application of sparse queries, and the emergent property with few iterations of training.
\item We offer three key insights to unlock the challenges of multi-dataset object detection training, including the calibration of the label space, the application and improvement of sparse queries, and the emergent property with few-iteration dataset-specific training.
\item Building upon these three insights, we introduce a simple yet flexible multi-dataset detection framework, denoted as Plain-Det, which satisfies the following criteria: flexibility to accommodate new datasets, robustness in performance across diverse datasets, training efficiency, and compatibility with various detection architectures.
\item We integrate Plain-Det into the Def-DETR model and conduct joint training on common public datasets, comprising 2,249 categories and 4 million images. This integration boosts the performance of the Def-DETR model from 46.9\% mAP on COCO to 51.9\%, achieving performance on par with the current state-of-the-art object detectors. Furthermore, it establishes new state-of-the-art results on multiple downstream datasets.
\end{itemize}

% \begin{figure}
%   \centering
%   \includegraphics[width=0.5\textwidth]{fig/fig2_LE.png} % 插入位于 "fig" 子目录中的图片
%   \caption{The left image is the similarity matrix of the regular classification head, the middle and right images represent the similarity matrices of the label embedding classification head before and after modification.}
%   \label{fig:fig3}
% \end{figure}

% \begin{figure}[t]
%     \centering
%     \caption{XXXX.} 
% % \includegraphics[width=0.5\textwidth]{fig/combined_fig2.pdf} 
%     \includesvg[width=0.9\textwidth]{fig/Fig2.svg}
%     \vspace{-2mm}
%     %要申明这个波动的训练曲线已经combine了xxx和xxx sampling策略。combine the class-aware repeat factor sampling strategy~\TODO{cite} and multi-dataset sampling strategy~\TODO{cite} to better balance class distributions and dataset sizes
% % \label{fig:fig2}
% \vspace{-6mm}
% \end{figure}

\section{Related work}
\label{sec:rw}

\textbf{Multi-Dataset Object Detection} 
is gaining prominence as a crucial step for advancing large vision models. It aims to break down the barriers between datasets and empower detection models with extensive data integration.
The fundamental challenges arise from the incongruities in dataset distributions, various taxonomies, and disparate data scales.
Early works~\cite{lambert2020mseg,zhao2021joining,zhao2020object} leverage external priors or human knowledge to construct a unified label space during multi-dataset training. However, this approach becomes inflexible when scaling dataset categories to thousands.
Recently, with the rise of joint image-text training, Many works~\cite{wang2023detecting, scaledet, detectionhub,vild,shi2023edadet} generate label spaces using text embeddings from pre-trained vision-language models.
However, due to the imbalanced data distribution in pre-trained text corpora, the text embedding space exhibits biases towards specific datasets. We are the first to identify and propose a solution to address this issue during the multi-dataset training process.

\noindent\textbf{Query-based Object Detection} has rapidly evolved with the ascendancy of the query-based~\cite{detr,dndetr,zhu2020deformable,sparsercnn,dino,kamath2021mdetr} framework, deriving various query initialization strategies.
The pioneering work DETR~\cite{detr} proposes a sequence of learnable queries for capturing object-level information. Naive zero-initialization is adopted originally. Later works~\cite{zhu2020deformable,sparsercnn,liu2022dabdetr,chen2022groupdetr,deta}, including Deformable DETR~\cite{zhu2020deformable} and Sparse RCNN~\cite{sparsercnn}, follow such design, accompanied by pixel-initialized queries or sparse proposals.
% Inspired by the two-stage RCNN~\cite{he2017mask}
% , ~\cite{zhu2020deformable} explores to introduce the spatial priors based on Deformable DETR. Specifically, the two-stage Deformable DETR~\cite{zhu2020deformable} reuses the generated region proposal in the first stage to propose the queries in the second stage. Recently, 
% DINO~\cite{dino} integrates previous strategies and introduces a mixed query selection, which initializes positional queries with proposals but leaves content queries learnable. Both studies showcase the remarkable effectiveness of the spatial prior in object detection.
In the context of multi-dataset training, Det-Hub~\cite{detectionhub} introduces dataset-aware queries to adapt to various data distributions. 
% However, such language-enhanced query lacks spatial priors, resulting in unsatisfactory performance in the object detection task.
To effectively integrate the spatial and dataset priors into our model, we meticulously devised the Class-Aware Query Compositor, initialing class-aware queries with weak priors.

\noindent\textbf{Data Sampling Strategy} holds critical importance in multi-dataset training~\cite{unidet,scaledet,detectionhub} due to substantial variations in the number of images and classes across different datasets. For instance, LVIS~\cite{gupta2019lvis} encompasses nearly four times the number of categories present in Objects365~\cite{shao2019objects365}, while the number of images is 17 times fewer. UniDet~\cite{unidet} has demonstrated that incorporating a uniform sampling strategy for various datasets yields benefits. However, such ``balance'' between different datasets solely refers to the sampled number of images.
% In this work, we believe that considering the training difficulty for each dataset is also essential since we should avoid repetitively sampling a well-trained dataset. 
Therefore, we propose a hardness-indicated sampling strategy to dynamically adjust sampler weights based on the training hardness of different datasets.

\section{Our Method}

\label{sec:method}
\begin{figure*}[t]
    \centering
    \caption{\textbf{Method overview.} Our multi-dataset detector Plain-Det is compatible with various query-based detection families. (a) Our multi-dataset joint training framework for object detection. 
    % : Each dataset possesses its own unique online sampler, query compositor, and classification head. Different datasets also share a semantic space through pre-trained VLM. 
    (b) Overview of query compositor: it takes images and the label embeddings of datasets as inputs and outputs class-aware query.}
\includegraphics[width=0.99\textwidth]{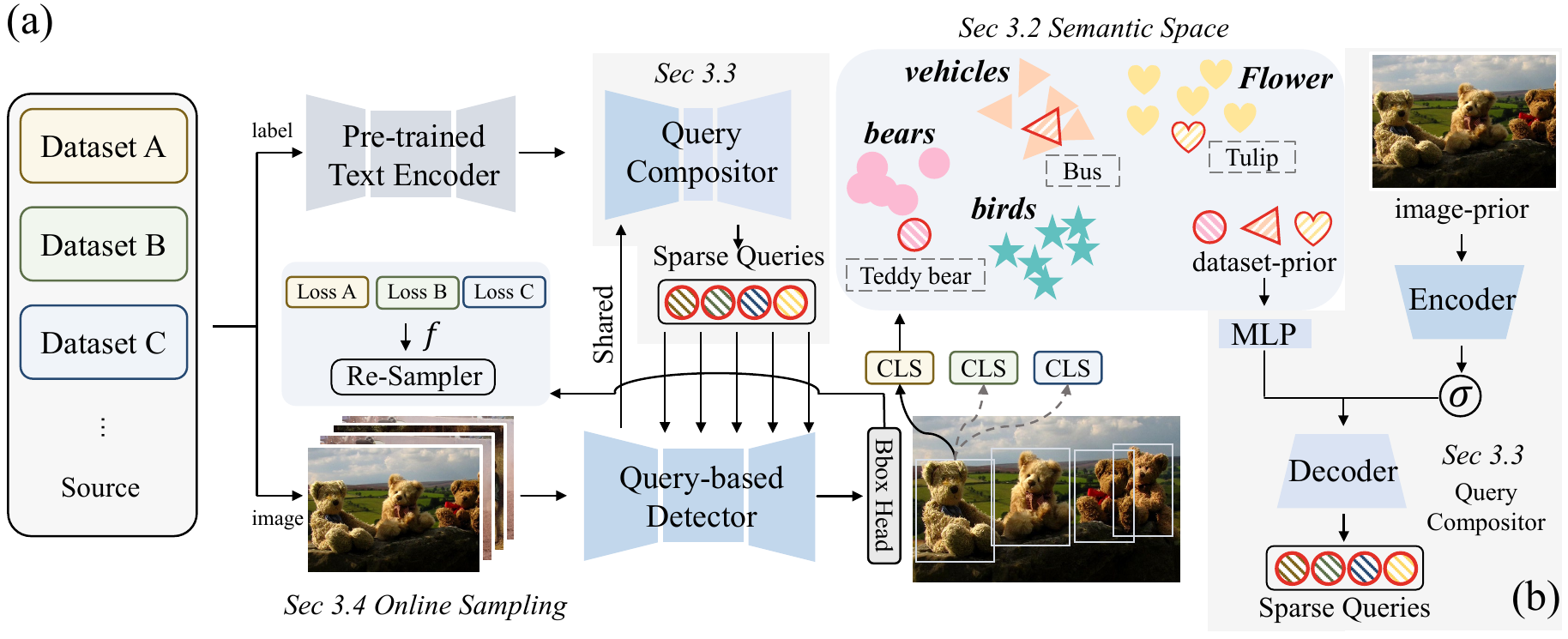} 
    \vspace{-2mm}
\label{fig:framrwork} 
\end{figure*}

As shown in Fig~\ref{fig:framrwork}, we propose a general framework and training strategy for multi-dataset object detection, free from pursuing a particular detection architecture, and compatible with various detection families.  
% This implies that our framework is compatible with various detection families, such as RCNN or DETR, as long as they adhere to a query-based approach, with only minor adjustments required for the detectors.
%In Sec~\ref{subsec:pre}, we first introduce the preliminaries on query-based object detectors and then abstract three primary components of multi-dataset object detection training: 
%1) the object detector architecture with dataset-specific classification heads and frozen classifier in Section~\ref{subsec:Label}, 2) object queries generation via the class-aware query compositor in Section~\ref{subsec:Query} and 3) the hardness-indicated sampler for multi-dataset training (Section~\ref{subsec:uncertainty}).
In Sec~\ref{subsec:pre}, we first introduce the preliminaries on query-based object detectors and then abstract three primary components of multi-dataset object detection: 
1) the object detector architecture with dataset-specific classification heads and frozen classifiers (see Section~\ref{subsec:Label}), 2) object queries generation via the class-aware query compositor (see Section~\ref{subsec:Query}) and 3) the hardness-indicated sampler for multi-dataset training (see Section~\ref{subsec:uncertainty}).
% how to construct label space, how to choose detector framework, and how to sample data.
% In the upcoming sections, we offer an extensive elucidation of each aspect in , Section~\ref{subsec:Query}, and Section~\ref{subsec:uncertainty}.

\subsection{Preliminaries}
\label{subsec:pre}
\textbf{Query-based object detector.} %Through a set of learnable queries, query-based approaches~\TODO{cite} achieve sparse object output, thus circumventing the need for pre-defined anchor boxes or hand-crafted post-processing procedures. 
By reformulating object detection as a set prediction problem~\cite{detr}, recent query-based object detectors, such as DETR-based methods~\cite{detr, zhu2020deformable, dino, dndetr} or the Sparse-RCNN~\cite{sparsercnn}, leverage learnable or dynamically selected object queries to directly generate predictions for the final object set. This approach eliminates the requirement for hand-crafted components like anchor presets~\cite{girshick2015fast, ren2015faster} and post-processing NMS~\cite{girshick2015fast, ren2015faster}. 
A query-based detector consists of three components: a set of object queries, an image encoder (\eg, Transformer encoder~\cite{attention} in DETR or CNN~\cite{cnn, resnet} in Sparse-RCNN), and a decoder (\eg, Transformer decoder~\cite{attention} in DETR or dynamic head~\cite{chen2020dynamic} in Sparse-RCNN). 
For a given image $I$, the image encoder $Enc(\cdot)$ extracts the image features, which are subsequently input to the decoder $Dec(\cdot)$ along with the object queries $\mathcal{Q}$ to predict the category $C$ and bounding box $B$ for each query. Typically, the classification head $\mathcal{H}_c(\cdot)$ and the box regression head $\mathcal{H}_b(\cdot)$ consist of several layers of multi-layer perceptrons (MLPs). The overall detection pipeline can be viewed as follows:
\begin{equation}
    \begin{aligned}
        \hat{\mathcal{Q}} &= Dec( Enc(I), \mathcal{Q}), \\
        C &= \mathcal{H}_c(\hat{\mathcal{Q}}), \quad B=\mathcal{H}_b(\hat{\mathcal{Q}}),
    \end{aligned}
\end{equation}
where $\hat{\mathcal{Q}}$ is the query feature after query refinement by decoder layer. To simplify the demonstration, we use $f(\cdot)$ to represent the object detector, where the encoder $Enc(\cdot)$, decoder $Dec(\cdot)$, learnable or selected queries $\mathcal{Q}$, classification head $\mathcal{H}_c(\cdot)$, and box regression head $\mathcal{H}_b(\cdot)$ are components of it. 

%Our method compiles with any query-based methods without requiring any alterations to the network architecture. However, to adapt to multiple datasets, specific classification heads, and queries have been introduced, as elaborated in Section~\ref{subsec:Label} and Section~\ref{subsec:Query}, respectively.
% \TODO{sc: Given an image $I$, the image encoder $E(\cdot)$ extracts the image features, which are fed to the decoder with the object queries $Q$ to predict xxx with regress class $C$ and bounding box $B$.} %定义出怎么预测的公式，定义出最后的分类头，后面公式要用。

%然后重点提出，我们compile with query-based xxx, 不需要更改任何的网络架构，但为了adapt to多数据集，有特别提出的分类头和query, 在xxx, xxx 介绍。

\noindent\textbf{Single-dataset object detection training.}
%这一章节主要讲training, 不讲架构。
%For 单数据集训练，定义单数据集的学习目标，参考Unidet公式[1]
% 然后直接讲多数据集，定义好数据集的符号等等。[3]
For training on a single dataset $D$, the optimization objective of a query-based object detector can be formulated as follows,
\begin{equation}
\underset{\Theta = \{Enc, Dec, \mathcal{Q}, \mathcal{H}_b,\mathcal{H}_c\}}{\operatorname{argmin}}
\quad
 \mathbb{E}_{(I,\hat{B}) \sim D}[\ell(f(I; \Theta), \hat{B})],
\end{equation}
where ($I$, $\hat{B}$) represents a pair of image and annotations from the dataset $D$. The loss function $\ell$ is typically the cross-entropy loss for the class predition and the generalized intersection over union loss for the box regression~\cite{giou}. 
% Different from single dataset object detection, multi-dataset training involves the consideration of M datasets, each with its distinct label space. Distinct datasets exhibit varying data distributions, and the presence of domain gaps contributes to differences in their learning complexities. For instance, COCO~\TODO{cite} and LVIS~\TODO{cite} both include approximately 10 million images from the same domain. However, COCO annotates only 80 common objects, whereas LVIS covers 1203 categories. On the other hand, Object365~\TODO{cite} comprises 365 categories, significantly fewer than LVIS, yet it boasts nearly ten times the amount of image data. The natural divergence in training data across different datasets leads us to address three fundamental questions. 
% \textcolor{violet}{\ding{182}} How can we ensure consistent training and efficiently leverage the advantages of distinct datasets, considering the combination of numerous datasets and a large number of labels? \textcolor{violet}{\ding{183}} How should we go about selecting and enhancing a detector based on the varying performance of different types of detectors when confronted with large-scale and complex distribution datasets?
% \textcolor{violet}{\ding{184}} How should we address the issue of imbalanced dataset sizes, data distribution, and training difficulty between various datasets? 
% We will address each of these questions in the upcoming sections. 

\subsection{Dataset-specific Head with Frozen Classifier}
\label{subsec:Label}
% label 拼接 -》 多头
% text embed 干扰

% 架构 with prediction head
% classifier
In this section, we introduce our multi-dataset object detection framework, which is compatible with any query-based object detection architecture. To support multiple datasets, our framework features a distinct dataset-specific classification head for each dataset. Within these heads, the classifiers are pre-extracted and frozen during training.

\noindent\textbf{Object detector with dataset-specific classification head.} Multiple datasets $D_1$, $D_2$, ..., $D_M$ with respective label spaces $L_1$, $L_2$, $...$, $L_M$, may have inconsistent taxonomies. For example, the \texttt{``dolphin''} class in the Obj365 dataset~\cite{shao2019objects365} is labeled as background in the COCO dataset~\cite{coco}. 
Recent works thus either manually or automatically create a unified label space for the $M$ datasets by concatenating every dataset-specific label space~\cite{scaledet}, learning mappings from each label space to the unified one~\cite{unidet}, or assigning soft labels to the sub-word set~\cite{detectionhub} of class names. However, a unified label space lacks flexibility when scaling to more datasets and tends to become noisier as the label space size grows. 
Therefore, we propose to keep each label space separate to directly and naturally address the issue of inconsistent taxonomies. 
Specifically, we augment the query-based object detector by adding $M$ dataset-specific classification heads $\mathcal{H}_{c}^{1}(\cdot)$, $\mathcal{H}_{c}^{2}(\cdot)$, ..., $\mathcal{H}_{c}^{M}(\cdot)$, each specializing in classifying objects within its corresponding label space:
\begin{equation}
    \begin{aligned}
        \hat{\mathcal{Q}} &= Dec( Enc(I), \mathcal{Q}), \\
        C^m &= \mathcal{H}_{c}^{m}(\hat{\mathcal{Q}}), \quad B=\mathcal{H}_b(\hat{\mathcal{Q}}), 
    \end{aligned}
\end{equation}
where $\mathcal{H}_{c}^{m}(\cdot)$ is the dataset $D_m$'s classification head on the label space $L_m$. The encoder $Enc(\cdot)$, decoder $Dec(\cdot)$, object queries $\mathcal{Q}$, and class-agnostic box regression head $\mathcal{H}_b(\hat{\mathcal{Q}})$ are shared across datasets. 
Notably, while our detector is formally similar to the partitioned detector~\cite{unidet}, our classification heads are independently optimized with their respective objectives. In contrast, partitioned detector~\cite{unidet} subsequently optimizes the outputs of the partitioned detector with the objective of unified taxonomy. 

%Next, we consider the methodology for calculating class losses among diverse datasets in a multi-dataset configuration. One possible solution is to directly concatenate the data. Still, as the size of the datasets increases, the total number of classes could become enormous and difficult to optimize (For instance, when we concatenate datasets like COCO, Object365, and LVIS, the number of classes swells to nearly two thousand.). Following UniDet~\TODO{cite}, we adopt a partitioned head as shown in Fig~\TODO{cite}. 

\noindent\textbf{Frozen classifiers with a shared semantic space.} 
While dataset-specific heads address conflicts arising from inconsistent taxonomies, they do not fully leverage similar semantic classes, such as the common class \texttt{``person''}, from different datasets for comprehensive learning. 
%This implies that during training, the labels from one dataset are invisible to the other dataset. 
To address this problem and transfer the common knowledge across diverse datasets, we follow~\cite{scaledet} to leverage the pre-trained CLIP~\cite{clip} model's feature space as the shared semantic space for class labels.
Specifically, for each dataset $D_m$ with label space $L_m$, we utilize its labels' CLIP text embeddings as the classifier $W^m$ within its classification head $\mathcal{H}_{c}^{m}(\cdot)$:
\begin{equation}
    W^m = Enc_{\text{text}}(\text{Prompt}(L_m)),
\end{equation}
where $\text{Prompt}(L_m)$ generates text prompts ``$\textit{the photo is}$ [class name]'' for every class in label space $L_m$, and $Enc_{\text{text}}(\cdot)$ is the frozen text encoder of CLIP. 
% Figure~\ref{fig:fig2} visualizes the similarities between text embeddings, showing that the semantic relevance among most classes adheres to visual or semantic associations we perceived and expected. 
% %However, there are instances where the similarity between certain categories and their coupling to all other categories is exceptionally high, such as ``person.'' 
% However, certain classes are exceptionally similar to others, such as \texttt{``person''}, which affects the classifier's discriminative ability. We suggest this could be related to data distribution during CLIP's pre-training, where words that frequently co-occur exhibit high similarity but may not reflect their semantic associations. This leads to the learned text encoder with inherent bias. For instance, \texttt{``person''} might be overly prevalent in the pre-training data. 
To correct the bias comes from CLIP's training data distribution, We calibrate text embeddings by removing the basis bias as follows:
% Given that the bias comes from CLIP's training data distribution, an empty string could be the basis to extract the biases in text embedding. We calibrate text embeddings by removing the basis bias as follows:
\begin{equation}
    \hat{W}^{m} =\text{Norm}(W^m -  \text{Enc}_{\text{text}}(\texttt{NULL})),
\end{equation}
where $\text{Enc}_{\text{text}}(\texttt{NULL})$ is the text embedding of the empty string, and the Norm is the L2 normalization. %As shown in Fig~\TODO{cite}, after regularization, we position our label space within a more suitable range.
\begin{figure*}[t]
\captionof{figure}{\textbf{Comparison of different proposal generation methods and ours.} (a) Proposal generation from sparse queries. (b) Proposal generation from top-K dataset-specific pixel features in dense image feature map. (c) Our class-aware query generation relies on weak priors associated with the dataset and the image. Dataset-specific head shows we use the different frozen classification heads to calculate the loss. 
}
% \scriptsize
\begin{tabular}{cc}
\fig[.96]{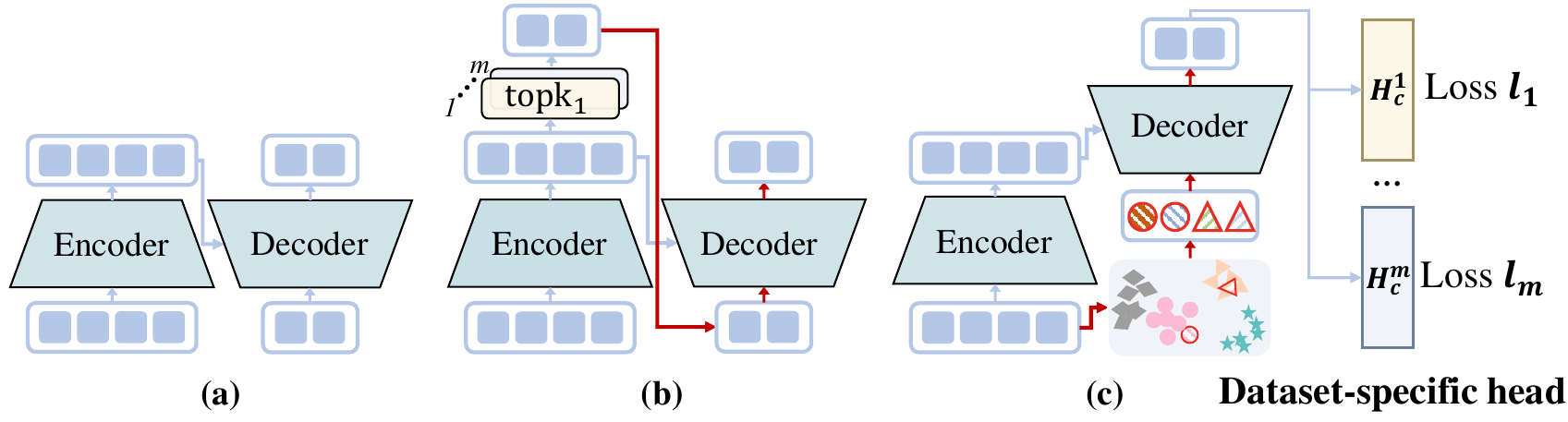} &
% \fig[.49]{fig/fig4_p2.pdf}\\

% \makecell{(a) Sampled Images and Features\\ in  \textcolor{coco}{COCO} \& \textcolor{obj}{Obj365}} &
% \makecell{(c) The performance-data volume \\ trade-off curve.}\\

\end{tabular}%
\label{fig:fig4}
% \par\vspace{24pt}
\end{figure*}

\subsection{Class-Aware Query Compositor}
\label{subsec:Query}
% How to choose a detector? How do you think you could improve your detector?
% 发现 init query 和 类别相关  
%%% check dino
\noindent\textbf{Object query generation}, as the essential component of query-based object detectors~\cite{sparsercnn,detr,dndetr,liu2022dabdetr,zhu2020deformable,chen2022groupdetr,deta,dino}, has been extensively explored in single-dataset training, yielding diverse types based on their independence from images~\cite{dino}. %, including image-agnostic static, learnable queries~\TODO{cite}, dynamic queries selected from the image~\TODO{cite}, and mixed queries by combining static content queries with dynamic anchors~\TODO{cite}. 
%In multi-dataset object detection, the initialization of object queries becomes even more crucial due to the additional challenge posed by the diversity of multiple datasets involved, extending beyond the consideration of images alone. 
%In multi-dataset object detection, the initialization of object queries becomes even more crucial due to the diversity of multiple datasets involved, extending beyond the scope of query initialization in single-dataset object detection, which solely considers the input image.
%In multi-dataset object detection, initializing object queries becomes even more crucial due to the diversity of multiple datasets involved. This extends beyond the scope of query initialization in single-dataset object detection, where queries are initialized randomly or generated solely based on the input image. 
In multi-dataset object detection, initializing object queries becomes even more crucial due to the diversity of multiple datasets involved. This extends beyond the scope of query initialization in single-dataset object detection, \textcolor{black}{where queries are initialized randomly or generated from input image feature map based on dataset-specific top-K score} (see Fig~\ref{fig:fig4}\textcolor{red}{a} and \textcolor{red}{b}).
In our preliminary experiments on multi-dataset object detection, selecting the top-K \textcolor{black}{pixel features}~\cite{detr} from the encoder (Def-DETR ++~\cite{zhu2020deformable} in Fig~\ref{fig:fig2}) led to a significant drop in performance, %, with a decrease of \TODO{num} AP on the LVIS dataset 
compared to single-dataset training. This is because the top-K candidate objects within an image depend significantly on the dataset taxonomy and are strongly correlated with the dataset. An excessively strong dataset prior conditions the detector towards dataset-specific decoding, which in turn hinders the decoder from fully exploiting multiple datasets for comprehensive learning. 
Conversely, dataset-agnostic query initialization (Def-DETR~\cite{zhu2020deformable} in Fig~\ref{fig:fig2}) shares the same learnable object queries across all datasets. %While this approach improves performance slightly compared to the previous initialization method, it still falls significantly short of the single-dataset performance. 
%%%%%%%%%%%%%%%%%%%%%%%%%%%%%%%%%%%%%%%%%%%
% However, learning a shared query set for multiple datasets with significantly different data distributions presents an exceedingly difficult challenge. This is because the learned query set needs to capture the distribution of categories, the distribution of counts, and the spatial distribution within each dataset.
%%%%%%%%%%%%%%%%%%%%%%%%%%%%%%%%%%%%%%%%%%%
%As shown in Anchor-Detr~\TODO{cite} and Conditional-Detr~\TODO{cite}, a context query often contains prior knowledge regarding the location of objects. 
%Based on \TODO{Fig}, we propose that in multi-dataset training, queries play an additional role in balancing the distribution disparities among datasets. 
Based on these observations and insights, we propose a novel query initialization method for multi-dataset object detection (see Fig~\ref{fig:fig4}\textcolor{red}{c}). Neither dataset-agnostic nor strongly dataset-dependent, our class-aware query initialization relies on weak priors associated with the dataset and the image. 
Given an image $I$ and its corresponding dataset $D_m$'s classifier $\hat{W}^m$, we first construct a dataset-specific weak query embedding $\mathcal{Q}^b$ based on the classifier as follows:
\begin{equation}
    \mathcal{Q}^b = \text{MLP} (\hat{W}^m).
\end{equation}
Notably, despite being dataset-specific, unlike strong priors that directly select dataset-specific image content, we obtain a weak prior by employing its dataset-specific classifier that shares the same semantic space across different datasets. Through this weak prior, similar semantic labels across different datasets can be shared. %尽管是dataset-specific的，但是不同于基于dataset直接选择image上的content 的strong prior, 我们用在不同数据集中share的semantic space上的classifier来获取prior是weak prior, 不同数据集间similar semantic labels通过这个weak prior也能共享的。
Subsequently, we opt to extract the global image feature rather than the top-K local content features as the weak image prior. We then combine this with dataset-specific queries as follows:
\begin{equation}
\label{equ: ini_query}
\begin{aligned}
    \mathcal{W} &= \text{MLP} (\text{Max-Pool}(Enc(I))), \\
    \mathcal{Q}^c &= \mathcal{W} \mathcal{Q}^b,
\end{aligned}
\end{equation}
where Max-Pool performs max pooling over the entire image, and $\mathcal{W}$ can be regarded as a weak image prior. And $\mathcal{Q}^c$ represents the final query feature fed to the decoder.
 %就定义下Max-Pool, W是image prior， Qc是最终的class-aware object quries. 
%Following this, composite the class-aware query by:
%\begin{equation}
%    \mathcal{Q}^c = \mathcal{W} \mathcal{Q}^b,
%\end{equation}
%As shown above, 
%Our modifications are solely related to the query initialization, thereby compatible with all the query-based object detection methods by modifying their query initialization. % to class-aware using this approach.
Importantly, our modifications are solely focused on the classification head and query initialization, making them easily applied to and compatible with all query-based object detectors.

\subsection{Training with Hardness-indicated Sampling}
\label{subsec:uncertainty}

Apart from detector architecture adjustments discussed in Sec~\ref{subsec:Label} and Sec~\ref{subsec:Query} to adapt to multiple datasets, training a multi-dataset detector brings forth additional challenges due to notable disparities in dataset distribution, image quantities, label space sizes, and more. In this section, we first formulate the objective of multi-dataset training and then improve the training strategy based on our observation of the emergent property introduced in Fig~\ref{fig:fig2}.

In general, the optimization objective for training our multi-dataset object detector on $M$ datasets $D_1$, $D_2$, ..., $D_M$ can be formulated as follows, 
\begin{equation}
\underset{\Theta \in \{Enc, Dec, \mathcal{Q}_c, \mathcal{H}_b\}, \{\mathcal{H}_c^{m}\}_{m=1}^M}
{\operatorname{argmin}} \mathbb{E}_{D_m} \big[
 \mathbb{E}_{(I,\hat{B}) \sim D_m}[\ell_m(f(I, \Theta; \mathcal{H}_c^{m}) , \hat{B})]\big],
\end{equation}
Where, except for the task-specific classification head $\mathcal{H}_c^{m}$, the remaining components of the object detector $f(\cdot)$, including the encoder $Enc(\cdot)$, decoder $Dec(\cdot)$, lightweight MLPs for generating object queries $\mathcal{Q}^c$ (Equ~\ref{equ: ini_query}), and class-agnostic box regression head $\mathcal{H}_b(\cdot)$, are shared across different datasets. Thanks to our dataset-specific classification head $\mathcal{H}_c^{m}(\cdot)$, our loss $\ell_m$ can be naturally tailored to the specific dataset, ensuring the preservation of the original training loss and sampling strategy for each dataset individually. For instance, we apply RFS~\cite{gupta2019lvis} to the long-tailed LVIS~\cite{gupta2019lvis} dataset but not to the COCO~\cite{coco} dataset.
%where different datasets share the same encoder-decoder $f$, basis query feature $\mathcal{Q}$ and class-agnostic $\mathcal{H}_b$, but they independently optimize their respective class heads $\mathcal{H}_{m}$ and $m$ $\in$ \{1...M\}.
%Thanks我们单独的classification head, the loss可以被formulated as unidet (4)参考，但是那个M_k我们要写成 f(; head_k)的样子，不是M个检测器。 然后就可以在D_k~{xxx}那里定义对数据集的sampling了。

Although dataset-specific losses can adapt to each dataset's internal characteristics, significant differences between datasets, such as differences in dataset sizes, present training challenges that must be addressed. 
%Since different datasets adhere to different data distributions $D_m$, when there is a large difference across datasets, the optimization process for training with many datasets might easily degenerate by focusing solely on optimizing one dataset, as shown in Fig~\ref{fig:fig2}. This might happen because one dataset is very complex or has a lot of data. To avoid the degeneration issue, 
Therefore, we propose a hardness-indicated sampling strategy to balance the number of images across datasets and dynamically assess dataset difficulty during online training.
% \textcolor{blue}{to resample the data distribution $D$ to rebalance the expected loss $\sum_m
%  \mathbb{E}_{(I,\hat{B}) \sim D_m}$.}  %不太对，平衡数据集的图像数量并 online判断数据集的困难程度
We first periodically record the box loss, $ L_1, \ldots, L_m $, for different datasets. Then compute the online sampling weight \( w_m \) as follows:
% \begin{equation}
%     {w_{m}
%     }=1+[
%         \frac{\max\{S_{m}\}}{\{S_{m}\}}
%         (\frac{\{\ell_{m}\}}{\min\{\ell_{m}\}}-1)
%     ]^\alpha
% \end{equation}
\begin{equation}
    {w_{m}
    }=\frac{L_{m}}{\min(\{L_{i}\}_{i=1}^{m})}
    [\frac{\max(\{S_{i}\}_{i=1}^{m}))}{S_{m}}]^\frac{1}{2},
\end{equation}
where $S_i$ means the number of images in the \textit{i-th} dataset, and $w_m$ will be involved in controlling the weight of each dataset in data sampling. The online sampler will sample data from each dataset according to the proportion of its corresponding weight $ w_m $. %\textcolor{blue}{explain}

\section{Experiment}
\label{sec:experiment}
We conduct the following experiments to demonstrate the effectiveness of our Plain-Det. We first introduce our training setups in Section~\ref{sec:data_para}. 
Then we analyze Plain-Det with an increasing number of datasets in Sec~\ref{subsec:increasing_dataset}, report Plain-Det on various detection benchmarks in Sec~\ref{subsec:multi-dataset} and Sec~\ref{subsec:coco-sota} and conduct ablation study in Sec~\ref{subsec:abla}.

% Then we evaluate the flexibility and robustness of our model with progressively expanding set of datasets in Section 4.2. Our model outperforms the state-of-the-art (SOTA) multi-dataset method~\cite{scaledet} and SOTA COCO~\cite{coco} detector consistently and significantly, as shown in Section 4.3 and 4.4. We further provide an ablation study in Section 4.5.

\subsection{Training setups}
\label{sec:data_para}

\noindent\textbf{Implementation details.} If not mentioned, we use a partitioned (with dataset-specific head) Deformable-DETR\cite{zhu2020deformable} as our default object detector and set the number of queries to 300. 
Profiting from our model's compatibility with the detection family, we also integrate Plain-Det with other query-based detectors, Sparse R-CNN\cite{sparsercnn}, for experiments. 
Our implementation is based on the official implementation in Detectron2~\cite{wu2019detectron2} and Detrex~\cite{ren2023detrex}. 
%修改
% We keep the default hyper-parameters and use the standard data augmentation.
% For backbone, we utilize ImageNet\cite{deng2009imagenet} pre-trained ResNet-50\cite{resnet} in most experiments and Swin Transformers~\cite{liu2021swin} for model scalability experiments. 
% To build the unified label space, we use label embedding generated by the pre-trained CLIP~\cite{clip} model as the classifier for each dataset. Our ResNet-50~\cite{resnet} models are trained on 8 NVIDIA A100 with Adam Optimizer; more details can be seen in Supplementary. 

\subsection{Performance with growing number of datasets}
\label{subsec:increasing_dataset}

\begin{table}[t]
    \centering
        \caption{\textbf{Performance with growing number of datasets}. 
        We train our Plain-D$^{\text{N}}$et under a strict multi-dataset setting: when the number of datasets increases, the total training iteration remains the same.
        For example, we train O365~\cite{shao2019objects365} for 450k iterations separately, then we train `C+L+O' for 450k iterations too. \scratch: method w/o our Plain-D$^{\text{N}}$et, \pretrain: method w/ our Plain-D$^{\text{N}}$et. Single: single-dataset training, mAP is the mean AP across datasets. 
        }\label{table:increasing_multi}
        \tablestyle{4pt}{1}
        \begin{tabular}{lccccc}\toprule
        Method & COCO~\cite{coco} & LVIS~\cite{gupta2019lvis}  &O365~\cite{shao2019objects365} & OID~\cite{zareian2021open} &  mAP\\\midrule
        \scratch Single~\cite{zhu2020deformable} & 45.6 & 33.6 & 32.2 & 61.0 & 43.2\\
        \scratch C+L+O+D & 44.2 & 29.1 & 27.4 &59.4 &40.0\\\midrule
        \pretrain L & 37.2 & 33.3 &13.4 & 35.3 &29.8\\
        \pretrain L+C & 46.0 & 33.2 & 14.2 & 35.7 &32.3\\
        \pretrain L+C+O & 51.8 & 39.9 & 33.2 & 41.7 &41.7\\
        \rowcolor[HTML]{efefef} \pretrain L+C+O+D & 51.9 & 40.9 &33.3 & 63.4 &47.4\\
        % \rowcolor[HTML]{efefef} \pretrain C+L+O+D & 49.2 & 37.4 &31.6 & 74.2 &\cellcolor[HTML]{efefef}-\\
        \bottomrule
        \end{tabular}
\end{table}
\noindent\textbf{Strict multi-dataset training.} To demonstrate the effectiveness of our Plain-Det in multi-dataset training, showcasing how different datasets can mutually benefit from our approach, we test our method under the strict multi-dataset training setting. Under the strict multi-dataset training setting in Table~\ref{table:increasing_multi}, when the number of datasets increases, the total training iteration remains the same as the largest dataset among them. For example, we train O365~\cite{shao2019objects365} for 450k iterations separately, then we train ‘L+C+O’ for 450k iterations too. Note that generally, increasing training iterations improves the performance of detectors. However, in strict settings, the performance improvement of multi-dataset training can only come from the mutual assistance of different datasets, which is precisely our focus. An exception arises with the ‘L+C+O+D’ dataset, where severe underfitting is observed when training only on O365~\cite{shao2019objects365} iterations. Consequently, we double the training iterations.

\noindent\textbf{Increasing performance when increasing number of datasets.} We provide the single-dataset baseline by training official Deformable-DETR~\cite{zhu2020deformable} which has a similar single-dataset performance (33.6\% vs 33.3\% on LVIS~\cite{gupta2019lvis}). We also provide a simple baseline for merging multi-dataset training to demonstrate that improving performance across multiple datasets is not a trivial task.  Starting from training on a single dataset, LVIS, the performance on COCO~\cite{coco} improved from 37.2\% to 46.0\%, 51.8\%, and 51.9\%. The results of multi-dataset training also surpassed the performance of single-dataset training at 45.6\%. The mAP of multi-dataset training also increased from 29.8\% to 32.3\%, 41.7\%, 47.4\%. Moreover, compared to the multi-dataset detector baseline, Plain-Det gives much better performance, even though overall training iteration does not extend which means that the utilization of Plain-Det facilitated mutual assistance among different datasets, leading to the learning of a unified detector.

\subsection{Comparison to state-of-the-art multi-dataset detectors}
\label{subsec:multi-dataset}
\begin{table}[t]
    \caption{\textbf{Compare to state-of-the-art multi-dataset detectors.}}
    \label{table:sota_multi}
    \vspace{-0mm}
    \begin{minipage}{0.48\textwidth}
        \centering
        \subcaption{\textbf{Performance of different methods~\cite{zhou2022detecting, scaledet} on combinations of different datasets~\cite{gupta2019lvis,coco,russakovsky2015imagenet,shao2019objects365,oid}.} We report the AP on COCO (C), LVIS (L), and mean AP across two datasets.}
        \vspace{-2mm}
        \tablestyle{1.5pt}{1.08}
        \begin{tabular}{ll|ll|c}
\toprule
Model & Dataset(s) & C & L & mAP \\
\hline
Detic~\cite{zhou2022detecting}  & L,C & 43.9 & 33.0 & 38.4\\
Detic~\cite{zhou2022detecting} & L,C,IN21K & 42.4 & 35.4 & 38.9 \\ \hline
ScaleDet~\cite{scaledet} & L,C & 44.9 & 33.3 & 39.1 \\
ScaleDet~\cite{scaledet} & L,C,O365 & 47.0 & 36.5 & 41.7 \\
ScaleDet~\cite{scaledet} & L,C,O365,D & 47.1 & 36.8 & 41.9 \\ \hline
Ours & L,C & 46.0 & 33.3 & 40.0 \\
Ours & L,C,O365 & 51.8 & 39.9 & 45.9 \\
\rowcolor[HTML]{efefef}Ours & L,C,O365,D & 51.9 & 40.9 & 46.4 \\
\hline
\end{tabular}
        \label{tab:tab1_p1}
    \end{minipage}
    \hspace{1.5mm}
    \centering
    \begin{minipage}{0.48\textwidth}
    \subcaption{\textbf{Performance gap between joint training and individual training of different methods~\cite{unidet, scaledet} on the multi-dataset training.} We report the AP on COCO (C), Object365 (O365), and mean AP across two datasets.}
    \begin{minipage}{0.43\textwidth}
        \vspace{-2mm}
        \centering
        \tablestyle{1.5pt}{1.08}
        
\begin{tabular}{ll|ll|c}
\toprule
Model & Dataset(s) & C & O365 & mAP \\
\hline
\multirow{3}{*}{UniDet~\cite{unidet}}
 & single & 42.5 & 24.9 & 33.7 \\
 & multiple & 45.5 & 24.6 & 35.0 \\
 & $\Delta$ & \scriptsize{\textcolor{green}{+3.0}} & \scriptsize{\textcolor{green}{+0.3}} & \scriptsize{\textcolor{green}{+1.3}} \\ \hline
 
\multirow{3}{*}{ScaleDet~\cite{scaledet}}
 & single & 46.8 & 28.8 & 37.8 \\
 & multiple & 47.1 & 30.6 & 38.9 \\
 & $\Delta$ & \scriptsize{\textcolor{green}{+0.3}} & \scriptsize{\textcolor{green}{+1.8}} & \scriptsize{\textcolor{green}{+1.1}} \\ \hline

\multirow{3}{*}{Ours} & single & 45.6 & 30.0 & 37.8 \\
& multiple & 51.9 & 33.3 & 42.6 \\
\rowcolor[HTML]{efefef} & $\Delta$ & \scriptsize{\textcolor{green}{+6.3}} & \scriptsize{\textcolor{green}{+3.3}} & \scriptsize{\textcolor{green}{+4.8}} \\ \hline
 
\hline
\end{tabular}
        \label{tab:tab1_p2}
    \end{minipage}
    \end{minipage} 
\end{table}
In Table~\ref{table:sota_multi}, we compare Plain-Det against other state-of-the-art multi-dataset detectors~\cite{zhou2022detecting,unidet,scaledet}. In Table~\ref{tab:tab1_p1}, We provide the performance of different methods~\cite{zhou2022detecting, scaledet} on combinations of different datasets~\cite{gupta2019lvis,coco,russakovsky2015imagenet,shao2019objects365,oid} for LVIS~\cite{gupta2019lvis} (L) and COCO~\cite{coco} (C) to demonstrate the robustness of different methods to different datasets. In Table~\ref{tab:tab1_p2}, We present the performance gap between joint training and individual training of different methods~\cite{unidet, scaledet} on the same multi-dataset COCO~\cite{coco} (C), Object365~\cite{shao2019objects365} (O365) and OID~\cite{oid} (D) to demonstrate the benefits of different methods on the same multi-dataset.

\noindent\textbf{Performance on different multi-datasets.} Table~\ref{tab:tab1_p1} shows the comparison between Plain-Det and previous multi-dataset detectors~\cite{zhou2022detecting, scaledet} on combinations of different datasets. 
% Compared with Detic~\cite{zhou2022detecting} using LVIS, COCO, ImageNet21k~\cite{russakovsky2015imagenet} (IN21k) as the multi-dataset training set, our Plain-Det outperforms Detic by 2.1\% on COCO and 0.3 \% on LVIS. In Detic, different datasets are merged using WordNet synsets, while Plain-Det utilizes partitioned classification heads and text embeddings~\cite{clip} to avoid interference caused by merging. When training on the large-scale dataset, Detic employs IN21k as weak supervision, where IN21k contains 14M images compared to O365's 1.7M. Plain-Det surpasses Detic by 7\% in terms of mAP, even when using less data (Even though classification and detection datasets cannot be directly compared, based on \cite{he2019rethinking} that 1 image in COCO~\cite{coco} is roughly equivalent to 7 images in ImageNet~\cite{deng2009imagenet}, the data we use is still much smaller than Detic). 
Compared with ScaleDet~\cite{scaledet}, our Plain-Det consistently outperforms ScaleDet~\cite{scaledet} across different dataset combinations (in an apple-to-apple comparison). The improvement increases from 0.9\% for ‘L+C’ to 4.2\% for ‘L+C+O’, and further to 4.5\% for ‘L+C+O+D’, demonstrating that our method scales better to larger datasets compared to ScaleDet. This also indicates that the three proposed improvements are highly effective.

\noindent\textbf{Performance increment.} Table~\ref{tab:tab1_p2} presents the comparison between Plain-Det and previous multi-dataset detectors~\cite{unidet, scaledet}, showcasing the performance gap between joint training and individual training.
% All methods in Table~\ref{tab:tab1_p2} share the same multi-dataset training set ‘C+O+D’, where ‘single’ denotes single dataset training, ‘multiple’ represents multi-dataset training, and $\Delta$ indicates the increment between those two training settings. 
Compared with Unidet~\cite{unidet}, which learns a unified label space, Plain-Det achieves an absolute increase of 7.6\% (42.6\% vs. 35.0\%) and a relative growth of over 3.5\% (4.8\% vs. 1.3\%), considering the average AP of two datasets. Compared with ScaleDet~\cite{scaledet}, which also utilizes CLIP~\cite{clip} text embeddings but overlooks the issues within text embeddings, Plain-Det achieves an absolute increase of 5.9\% (42.6\% vs. 36.7\%) and a relative growth of over 2.4\% (4.8\% vs. 2.4\%), considering the average AP. 
% \clearpage
\subsection{Comparisons of performance and efficiency across different detection families}
\label{subsec:coco-sota}
% \begin{table}[t]
%     \caption{\textbf{Compare to COCO SOTA detectors.}}
%     \label{tab:sota_coco}
%     \vspace{-0mm}
%     \begin{minipage}{0.48\textwidth}
%         \centering
%         \subcaption{\textbf{Comparison under RN50 backbone}}
%         \vspace{-2mm}
%         \tablestyle{1.5pt}{1.08}
%         \input{tables/rn50_coco}
%         \label{tab:tab2_p1}
%     \end{minipage}
%     \hspace{1.5mm}
%     \centering
%     \begin{minipage}{0.48\textwidth}
%     \subcaption{\textbf{Comparison under large backbone}}
%     \begin{minipage}{0.43\textwidth}
%         \vspace{-2mm}
%         \centering
%         \tablestyle{1.5pt}{1.08}
%         \input{tables/vit_coco}
%         \label{tab:tab2_p2}
%     \end{minipage}
%     \end{minipage} 
% \end{table}

% \begin{table}[t]
%     \caption{\textbf{Compare to COCO SOTA detectors.}}

%     \centering
%     \tablestyle{1.5pt}{1.08}
%     \input{tables/rn50_coco}
%     \label{tab:tab2_p1}

% \end{table}

\begin{table}[t]
    \caption{\textbf{Comparisons of performance and efficiency across different detection families.} We present the results of various multi-dataset training methods on the COCO validation set, along with the total number of iterations where COCO data appeared during multi-dataset training. 
    % Additionally, we provide the single-dataset performance of the underlying object detector upon which it relies. Our Plain-D$^{\text{N}}$et is the sole approach that significantly improves performance and efficiency across different detection frameworks.
    }
    \label{tab:sota_coco}
    \vspace{-3mm}
    \begin{minipage}{0.47\textwidth}
        \centering
        \subcaption{\textbf{Comparison under non-RCNN family}}
        \vspace{-2mm}
        \tablestyle{0.5pt}{1.08}
        \begin{tabular}{lccc}\toprule
        \makecell{non-RCNN\\family}  & \makecell{multi\\datasets} & epoch & \makecell{\textcolor{gray}{COCO}\\ AP$^{\text{box}}$} \\\midrule
        CenterNet2~\cite{zhou2021probabilistic}  & \xmark & 12  &42.9\\ 
        +Detic\cite{zhou2022detecting}  & \cmark & - & 42.4 \\ 
        +ScaleDet~\cite{scaledet} & \cmark & 192 & 47.1 \\ \hline
        Def-DETR\cite{zhu2020deformable}   & \xmark & 50 & 46.9 \\ 
        % Def-DETR\cite{zhu2020deformable}   & \xmark & 300 & 45.5 \\ 
        \rowcolor[HTML]{efefef} + Ours & \cmark & 36 & 51.9 \\ 
        \bottomrule
        \end{tabular}
        \label{tab:tab3_p1}
    \end{minipage}
    \hspace{1.5mm}
    \centering
    \begin{minipage}{0.47\textwidth}
    \subcaption{\textbf{Comparison under RCNN family}}
    \begin{minipage}{0.43\textwidth}
        \vspace{-2mm}
        \centering
        \tablestyle{0.5pt}{1.08}
        \begin{tabular}{lccc}\toprule
        \makecell{RCNN\\family}  & \makecell{multi\\datasets} & epoch & \makecell{\textcolor{gray}{COCO}\\ AP$^{\text{box}}$} \\\midrule
        Mask RCNN~\cite{he2017mask} & \xmark & 36 &39.8 \\ 
        +RegionCLIP~\cite{zhong2021regionclip}  & \cmark & - &42.7 \\ \hline
        Sparse RCNN\cite{sparsercnn}   & \xmark & 12 & 43.0 \\ 
         + Det-Hub~\cite{detectionhub} & \cmark & 12 & 45.3 \\ 
        \rowcolor[HTML]{efefef} + Ours & \cmark & 12 & 46.1 \\ 
        \bottomrule
        \end{tabular}
        \label{tab:tab3_p2}
    \end{minipage}
    \end{minipage} 
    \vspace{-2mm}
\end{table}

\noindent\textbf{Comparison under non-RCNN family.} Table~\ref{tab:tab3_p1} shows the comparison between Plain-Det and previous multi-dataset detectors~\cite{zhou2022detecting, scaledet} based on non-RCNN detectors~\cite{zhou2021probabilistic, zhu2020deformable}. We report the number of epochs where COCO~\cite{coco} appears during multi-dataset training to represent the efficiency of the multi-dataset object detector. A lower frequency of COCO appearances indicates higher training efficiency. Compared with the previous SOTA method ScaleDet~\cite{scaledet}, Plain-Det achieves superior performance (51.9\% vs 47.1\%) while utilizing fewer instances of COCO data (36 vs 192). Compared to our underlying object detector Def-DETR~\cite{zhu2020deformable}, the advantages of multi-dataset training are also significant. We achieved a notable performance boost of 5.0\% in COCO validation AP while using only 72\% of the data.

\noindent\textbf{Comparison under RCNN family.} Table~\ref{tab:tab3_p2} presents the comparison between Plain-Det and previous multi-dataset detectors~\cite{zhong2021regionclip, detectionhub} based on RCNN detectors~\cite{he2017mask, sparsercnn}. Sparse RCNN, as the first work to introduce queries into the RCNN series, has shown significant improvements in both performance and efficiency. By integrating Plain-Det into Sparse RCNN and utilizing the same quantity of COCO data, we achieved a performance increase of 3.1\%, reaching a COCO validation performance of 46.1\%, matching the best multi-dataset object detector under the RCNN series, Det-Hub~\cite{detectionhub}.

\begin{table*}[t]
  \centering
  \caption{{\bf Results of zero-shot transfer on 5 individual datasets on ODinW.} 
  R, T: ResNet50~\cite{resnet}, Swin-Tiny~\cite{liu2021swin}. GoldG: 0.8M
 grounding data curated by MDETR~\cite{kamath2021mdetr}, cap4M: 4M image-text pairs~\cite{cc12m}.
  }
  \vspace{-3mm}
  \resizebox{0.8\linewidth}{!}{%
  \begin{tabular}{l | c | c | ccccc c}
    \toprule
    Model & Datasets & \#Data & \#1  & \#2 & \#3& \#4 & \#5  & \cellcolor[HTML]{efefef} mAP \\
    \midrule
    GLIP-T\cite{Li_2022_CVPR} & O+GoldG+Cap4M & 5.5M  & 18.4 & 50.0 & 49.6 & 57.8& 44.1  & \cellcolor[HTML]{efefef}44.0 \\
 
    ScaleDet-R~\cite{scaledet} & L+C & 0.2M  & 9.1 & 2.1 &12.4 &41.2 &25.5 & \cellcolor[HTML]{efefef}18.1 \\
    ScaleDet-R~\cite{scaledet} & L+C+O & 1.9M  & 8.7 & 1.4 & 25.0 & 47.3 &20.8 & \cellcolor[HTML]{efefef}20.6 \\
    ScaleDet-R~\cite{scaledet} & L+C+O+D & 3.6M  & 23.2 & 45.1 & 38.9 & 48.3 &40.6 & \cellcolor[HTML]{efefef}39.2 \\
    \midrule 
    Ours-R & L+C & 0.2M  & 16.0 & 7.0 & 15.5& 45.9 & 26.6& \cellcolor[HTML]{efefef}22.2 \\
    Ours-R & L+C+O & 1.9M  & 16.5 & 13.3 & 31.4& 49.4 & 52.8& \cellcolor[HTML]{efefef}32.7 \\
    \rowcolor[HTML]{efefef} Ours-R & L+C+O+D & 3.6M  & 27.9 & 43.3 & 47.3& 58.1&54.1& \textbf{46.1} \\
    \bottomrule
  \end{tabular}
  }
  % \vskip 1em
  
  \label{tab:odinw4}
\end{table*}
\subsection{ODinW benchmark}
\label{subsec:odinw}

\noindent\textbf{Zero-shot transfer on ODinW.} Following ScaleDet~\cite{scaledet}, table~\ref{tab:odinw4} shows the zero-shot transfer performance on 5 individual datasets from Object Detection in Wild (ODinW~\cite{zhang2022glipv2}). 
% Consistent with our previous observations, detectors trained on more datasets tend to exhibit stronger generalization capabilities on out-of-domain data as the dataset expands. Our method consistently and steadily improves performance with increased datasets ( 22.2\% $\to$ 32.7\% $\to$ 46.1\%). 
For an apple-to-apple comparison, our method significantly surpasses the previous state-of-the-art ScaleDet~\cite{scaledet} in various training dimensions (+4.1\% in L+C, +12.1\% in L+C+O, +6.9\% in L+C+O+D). In comparison to GLIP~\cite{zhang2022glipv2}, which was trained on a much larger combination of multi-datasets, our method still outperforms GLIP with the use of fewer data (3.6M vs 5.5M), achieving a higher performance of 46.1\% compared to 44.0\%.

% \noindent\textbf{Fine-tune transfer on ODinW.} Table~\ref{tab:odinw13} represents the fine-tune tranfer performance on all 13 datasets in ODinW~\cite{zhang2022glipv2}. We fine-tune Plain-Det on each dataset with

\subsection{Ablation Study}
\label{subsec:abla}
\begin{table}[t]
    \caption{\textbf{Ablation on partition head, class-aware query and sampler.}}
    \label{tab:main2}
    \begin{minipage}{0.54\textwidth}
    \vspace{-7mm}
        \centering
        \subcaption{We ablate the partition head, sparse query on C+L, and sampler on C+L+O.}
        \vspace{-2mm}
        \tablestyle{1.5pt}{1.00}
        \renewcommand{\arraystretch}{0.95}
\begin{tabular}{lcccccc}
\toprule
 & \texttt{Par} & \texttt{Que} & \texttt{Tex} & C\cite{coco}&  L\cite{gupta2019lvis} & mAP \\  
% \hline 0  & \xmark & \xmark & \xmark & - & - & -\\
1  & \xmark & \xmark & \cmark & 39.3 & 23.8 & 31.6\\
2 & \cmark& \xmark & \cmark & 38.1 & 24.0 & 31.1 \\
3  & \cmark  & \cmark & \xmark & 44.2 & 28.7 &  36.5 \\
\rowcolor[HTML]{efefef}4 & \cmark  & \cmark & \cmark & \textbf{45.3} & \textbf{30.2} &\textbf{37.8} \\
\hline
 &\multicolumn{2}{c}{\texttt{Sampler}}   & C\cite{coco}&  L\cite{gupta2019lvis}  & O365~\cite{shao2019objects365}& mAP \\  
\hline 4  & \multicolumn{2}{c}{\xmark} &43.5  &26.8  &25.5  &31.9 \\
\rowcolor[HTML]{efefef}5 & \multicolumn{2}{c}{\cmark} & \textbf{47.1} &\textbf{32.4}  &\textbf{25.6}  &\textbf{35.0} \\
% \rowcolor[HTML]{efefef}4 & \cmark & \cmark & 43.4 & 28.1 & 34.8 \\
\hline
\end{tabular}
        \label{tab:tab4_p1}
    \end{minipage}
    \hspace{1.5mm}
    \centering
    \begin{minipage}{0.310\textwidth}
    \vspace{-1mm}
    \subcaption{Analysis on ratio $w$.}
    \begin{minipage}{0.27\textwidth}
        \vspace{-2mm}
        \centering
        % \tablestyle{1.5pt}{1.08}
        \fig[3.8]{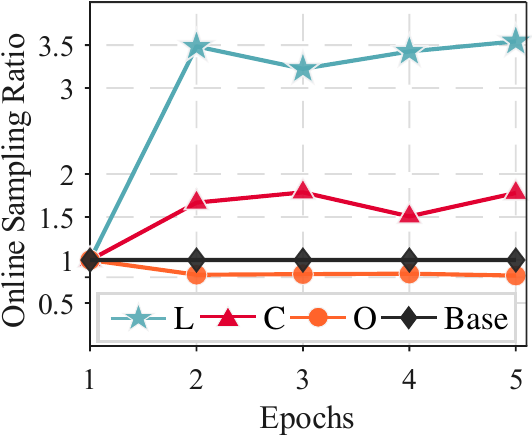}
        \input
        \label{tab:tab4_p2}
    \end{minipage}
    \end{minipage} 
    \vspace{-5mm}
\end{table}
% As shown in Table~\ref{tab:tab4_p1}, to evaluate the effectiveness of our algorithm, we conduct comprehensive and detailed ablation experiments on each component we propose.

\noindent\textbf{Ablation on partition head, sparse query, and label corr.} As we gradually introduce the components proposed, including \texttt{Que} (class-aware query) and \texttt{Tex} (label space correction), the mAP steadily increases from 31.1\% to 36.5\% and then to 37.8\%. One exception is the partition head. With the introduction of the partition head, there is a slight performance decrease (-0.5\%). However, we want to emphasize that a partition head is essential for the flexibility of scaling up datasets. Therefore, a minor performance loss is deemed acceptable.

\noindent\textbf{Ablation on sampler.} We conducted experiments on samplers using three datasets with highly uneven sizes (C+L+O), where L has the most categories and O has the most annotations and our online sampler results in a 4.1\% increase in mAP. As shown in Fig~\ref{tab:tab4_p2}, the online sampler oversampled from L and undersampled from O, which aligns with our prediction of the difficulty levels of the L and O datasets. 
% \subsection{Qualitative results}
% \label{subsec:qualitative}
% \label{sec:method}
% \begin{figure*}[t]
%     \centering
%     \caption{}
% \includegraphics[width=0.99\textwidth]{fig/fig6.pdf} 
%     % \vspace{-6mm} 
% \end{figure*}
\vspace{-5mm}
\section{Conclusion and Limitations}
\label{sec:conclusion}
We introduce \ours, an efficient multi-dataset detector, that meets the following requirements: adaptability to incorporate new datasets, consistency in performance across a wide range of datasets, efficiency in training, and compatibility with different detection architectures.
% effectively enhances in-domain performance and improves out-of-domain generalization capabilities through scaling up dataset sizes. In our paper, We introduce three critical factors influencing multi-dataset training. For each of these factors, we propose our own analysis and improvements. 
We validated our conclusions using over 17 datasets and achieved consistent improvements across multiple datasets. \noindent \textbf{Limitations:} Given that our semantic label space is derived from the CLIP~\cite{clip} models, we acknowledge that biases and controversies inherent in the training data for these models may be introduced into our model.
\noindent\textbf{Acknowledgment:} This work was supported by the National Natural Science Foundation of China (No.62206174) %the Shanghai Frontiers Science Center of Human-centered Artificial Intelligence (ShangHAI),
and MoE Key Laboratory of Intelligent Perception and Human-Machine Collaboration (ShanghaiTech University).

% \clearpage  % TODO REVIEW/FINAL: This \clearpage needs to be removed from both review and camera-ready versions.

\bibliographystyle{splncs04}
\bibliography{main}

\begin{thebibliography}{10}
\providecommand{\url}[1]{\texttt{#1}}
\providecommand{\urlprefix}{URL }
\providecommand{\doi}[1]{https://doi.org/#1}

\bibitem{detr}
Carion, N., Massa, F., Synnaeve, G., Usunier, N., Kirillov, A., Zagoruyko, S.: End-to-end object detection with transformers. In: Computer Vision--ECCV 2020: 16th European Conference, Glasgow, UK, August 23--28, 2020, Proceedings, Part I 16. pp. 213--229. Springer (2020)

\bibitem{cc12m}
Changpinyo, S., Sharma, P., Ding, N., Soricut, R.: Conceptual 12m: Pushing web-scale image-text pre-training to recognize long-tail visual concepts. In: Proceedings of the IEEE/CVF Conference on Computer Vision and Pattern Recognition. pp. 3558--3568 (2021)

\bibitem{chen2022groupdetr}
Chen, Q., Chen, X., Zeng, G., Wang, J.: Group detr: Fast training convergence with decoupled one-to-many label assignment. arXiv preprint arXiv:2207.13085  (2022)

\bibitem{scaledet}
Chen, Y., Wang, M., Mittal, A., Xu, Z., Favaro, P., Tighe, J., Modolo, D.: Scaledet: A scalable multi-dataset object detector. In: Proceedings of the IEEE/CVF Conference on Computer Vision and Pattern Recognition. pp. 7288--7297 (2023)

\bibitem{chen2020dynamic}
Chen, Y., Dai, X., Liu, M., Chen, D., Yuan, L., Liu, Z.: Dynamic convolution: Attention over convolution kernels. In: Proceedings of the IEEE/CVF conference on computer vision and pattern recognition. pp. 11030--11039 (2020)

\bibitem{deng2009imagenet}
Deng, J., Dong, W., Socher, R., Li, L.J., Li, K., Fei-Fei, L.: Imagenet: A large-scale hierarchical image database. In: 2009 IEEE conference on computer vision and pattern recognition. pp. 248--255. Ieee (2009)

\bibitem{girshick2015fast}
Girshick, R.: Fast r-cnn. In: Proceedings of the IEEE international conference on computer vision. pp. 1440--1448 (2015)

\bibitem{vild}
Gu, X., Lin, T.Y., Kuo, W., Cui, Y.: Open-vocabulary object detection via vision and language knowledge distillation. arXiv preprint arXiv:2104.13921  (2021)

\bibitem{gupta2019lvis}
Gupta, A., Dollar, P., Girshick, R.: {LVIS}: A dataset for large vocabulary instance segmentation. In: CVPR (2019)

\bibitem{he2017mask}
He, K., Gkioxari, G., Doll{\'a}r, P., Girshick, R.: {Mask R-CNN}. In: CVPR (2017)

\bibitem{resnet}
He, K., Zhang, X., Ren, S., Sun, J.: Deep residual learning for image recognition. In: CVPR (2016)

\bibitem{kamath2021mdetr}
Kamath, A., Singh, M., LeCun, Y., Synnaeve, G., Misra, I., Carion, N.: Mdetr-modulated detection for end-to-end multi-modal understanding. In: CVPR (2021)

\bibitem{sam}
Kirillov, A., Mintun, E., Ravi, N., Mao, H., Rolland, C., Gustafson, L., Xiao, T., Whitehead, S., Berg, A.C., Lo, W.Y., et~al.: Segment anything. arXiv preprint arXiv:2304.02643  (2023)

\bibitem{cnn}
Krizhevsky, A., Sutskever, I., Hinton, G.E.: Imagenet classification with deep convolutional neural networks. Advances in neural information processing systems  \textbf{25} (2012)

\bibitem{oid}
Kuznetsova, A., Rom, H., Alldrin, N., Uijlings, J., Krasin, I., Pont-Tuset, J., Kamali, S., Popov, S., Malloci, M., Kolesnikov, A., et~al.: The open images dataset v4: Unified image classification, object detection, and visual relationship detection at scale. International Journal of Computer Vision  \textbf{128}(7),  1956--1981 (2020)

\bibitem{lambert2020mseg}
Lambert, J., Liu, Z., Sener, O., Hays, J., Koltun, V.: Mseg: A composite dataset for multi-domain semantic segmentation. In: Proceedings of the IEEE/CVF conference on computer vision and pattern recognition. pp. 2879--2888 (2020)

\bibitem{dndetr}
Li, F., Zhang, H., Liu, S., Guo, J., Ni, L.M., Zhang, L.: Dn-detr: Accelerate detr training by introducing query denoising. In: Proceedings of the IEEE/CVF Conference on Computer Vision and Pattern Recognition. pp. 13619--13627 (2022)

\bibitem{Li_2022_CVPR}
Li, L.H., Zhang, P., Zhang, H., Yang, J., Li, C., Zhong, Y., Wang, L., Yuan, L., Zhang, L., Hwang, J.N., Chang, K.W., Gao, J.: Grounded language-image pre-training. In: Proceedings of the IEEE/CVF Conference on Computer Vision and Pattern Recognition (CVPR). pp. 10965--10975 (June 2022)

\bibitem{coco}
Lin, T.Y., Maire, M., Belongie, S., Hays, J., Perona, P., Ramanan, D., Doll{\'a}r, P., Zitnick, C.L.: Microsoft coco: Common objects in context. In: ECCV (2014)

\bibitem{liu2022dabdetr}
Liu, S., Li, F., Zhang, H., Yang, X., Qi, X., Su, H., Zhu, J., Zhang, L.: Dab-detr: Dynamic anchor boxes are better queries for detr. arXiv preprint arXiv:2201.12329  (2022)

\bibitem{liu2021swin}
Liu, Z., Lin, Y., Cao, Y., Hu, H., Wei, Y., Zhang, Z., Lin, S., Guo, B.: Swin transformer: Hierarchical vision transformer using shifted windows. In: Proceedings of the IEEE/CVF international conference on computer vision. pp. 10012--10022 (2021)

\bibitem{detectionhub}
Meng, L., Dai, X., Chen, Y., Zhang, P., Chen, D., Liu, M., Wang, J., Wu, Z., Yuan, L., Jiang, Y.G.: Detection hub: Unifying object detection datasets via query adaptation on language embedding. In: Proceedings of the IEEE/CVF Conference on Computer Vision and Pattern Recognition. pp. 11402--11411 (2023)

\bibitem{clip}
Radford, A., Kim, J.W., Hallacy, C., Ramesh, A., Goh, G., Agarwal, S., Sastry, G., Askell, A., Mishkin, P., Clark, J., et~al.: Learning transferable visual models from natural language supervision. In: International conference on machine learning. pp. 8748--8763. PMLR (2021)

\bibitem{redmon2017yolo9000}
Redmon, J., Farhadi, A.: Yolo9000: better, faster, stronger. In: CVPR (2017)

\bibitem{ren2015faster}
Ren, S., He, K., Girshick, R., Sun, J.: {Faster R-CNN: Towards Real-Time Object Detection with Region Proposal Networks}. NeurIPS  \textbf{28} (2015)

\bibitem{ren2023detrex}
Ren, T., Liu, S., Li, F., Zhang, H., Zeng, A., Yang, J., Liao, X., Jia, D., Li, H., Cao, H., Wang, J., Zeng, Z., Qi, X., Yuan, Y., Yang, J., Zhang, L.: detrex: Benchmarking detection transformers (2023)

\bibitem{giou}
Rezatofighi, H., Tsoi, N., Gwak, J., Sadeghian, A., Reid, I., Savarese, S.: Generalized intersection over union: A metric and a loss for bounding box regression. In: Proceedings of the IEEE/CVF conference on computer vision and pattern recognition. pp. 658--666 (2019)

\bibitem{russakovsky2015imagenet}
Russakovsky, O., Deng, J., Su, H., Krause, J., Satheesh, S., Ma, S., Huang, Z., Karpathy, A., Khosla, A., Bernstein, M., et~al.: Imagenet large scale visual recognition challenge. IJCV  \textbf{115}(3),  211--252 (2015)

\bibitem{shao2019objects365}
Shao, S., Li, Z., Zhang, T., Peng, C., Yu, G., Zhang, X., Li, J., Sun, J.: Objects365: A large-scale, high-quality dataset for object detection. In: ICCV (2019)

\bibitem{shi2023edadet}
Shi, C., Yang, S.: Edadet: Open-vocabulary object detection using early dense alignment. In: Proceedings of the IEEE/CVF International Conference on Computer Vision. pp. 15724--15734 (2023)

\bibitem{sparsercnn}
Sun, P., Zhang, R., Jiang, Y., Kong, T., Xu, C., Zhan, W., Tomizuka, M., Li, L., Yuan, Z., Wang, C., et~al.: Sparse r-cnn: End-to-end object detection with learnable proposals. In: Proceedings of the IEEE/CVF conference on computer vision and pattern recognition. pp. 14454--14463 (2021)

\bibitem{attention}
Vaswani, A., Shazeer, N., Parmar, N., Uszkoreit, J., Jones, L., Gomez, A.N., Kaiser, {\L}., Polosukhin, I.: Attention is all you need. Advances in neural information processing systems  \textbf{30} (2017)

\bibitem{wang2023detecting}
Wang, Z., Li, Y., Chen, X., Lim, S.N., Torralba, A., Zhao, H., Wang, S.: Detecting everything in the open world: Towards universal object detection. In: Proceedings of the IEEE/CVF Conference on Computer Vision and Pattern Recognition. pp. 11433--11443 (2023)

\bibitem{wu2019detectron2}
Wu, Y., Kirillov, A., Massa, F., Lo, W.Y., Girshick, R.: Detectron2. \url{https://github.com/facebookresearch/detectron2} (2019)

\bibitem{zareian2021open}
Zareian, A., Rosa, K.D., Hu, D.H., Chang, S.F.: Open-vocabulary object detection using captions. In: CVPR (2021)

\bibitem{dino}
Zhang, H., Li, F., Liu, S., Zhang, L., Su, H., Zhu, J., Ni, L.M., Shum, H.Y.: Dino: Detr with improved denoising anchor boxes for end-to-end object detection. arXiv preprint arXiv:2203.03605  (2022)

\bibitem{zhang2022glipv2}
Zhang, H., Zhang, P., Hu, X., Chen, Y.C., Li, L., Dai, X., Wang, L., Yuan, L., Hwang, J.N., Gao, J.: Glipv2: Unifying localization and vision-language understanding. Advances in Neural Information Processing Systems  \textbf{35},  36067--36080 (2022)

\bibitem{zhao2021joining}
Zhao, J., Ou, M., Xue, L., Cui, Y., Wu, S., Chen, G.: Joining datasets via data augmentation in the label space for neural networks (2021)

\bibitem{zhao2020object}
Zhao, X., Schulter, S., Sharma, G., Tsai, Y.H., Chandraker, M., Wu, Y.: Object detection with a unified label space from multiple datasets (2020)

\bibitem{zhong2021regionclip}
Zhong, Y., Yang, J., Zhang, P., Li, C., Codella, N., Li, L.H., Zhou, L., Dai, X., Yuan, L., Li, Y., et~al.: Regionclip: Region-based language-image pretraining. arXiv preprint arXiv:2112.09106  (2021)

\bibitem{zhou2022detecting}
Zhou, X., Girdhar, R., Joulin, A., Kr{\"a}henb{\"u}hl, P., Misra, I.: Detecting twenty-thousand classes using image-level supervision. arXiv preprint arXiv:2201.02605  (2022)

\bibitem{zhou2021probabilistic}
Zhou, X., Koltun, V., Kr{\"a}henb{\"u}hl, P.: Probabilistic two-stage detection. arXiv preprint arXiv:2103.07461  (2021)

\bibitem{unidet}
Zhou, X., Koltun, V., Kr{\"a}henb{\"u}hl, P.: Simple multi-dataset detection. In: Proceedings of the IEEE/CVF Conference on Computer Vision and Pattern Recognition. pp. 7571--7580 (2022)

\bibitem{zhu2020deformable}
Zhu, X., Su, W., Lu, L., Li, B., Wang, X., Dai, J.: Deformable detr: Deformable transformers for end-to-end object detection. arXiv preprint arXiv:2010.04159  (2020)

\bibitem{deta}
Zong, Z., Song, G., Liu, Y.: Detrs with collaborative hybrid assignments training. In: Proceedings of the IEEE/CVF international conference on computer vision. pp. 6748--6758 (2023)

\end{thebibliography}
\end{document}

\title{Supplementary Material for Plain-Det: \\ A Plain Multi-Dataset Object Detector} 


\appendix

In the Appendix, we provide additional information regarding,
\begin{itemize}
% [topsep=0pt,itemsep=-1ex,partopsep=1ex,parsep=1ex]
    \item Implementation Details (Appendix~\ref{appendix:impl_details})
    \item Detailed Results on ODinW (Appendix~\ref{sec:detailed-ODinW})
    \item Analysis on Proposal Generation (Appendix~\ref{sec:proposal-gen})
    \item Qualitative Results (Appendix~\ref{sec:qualitative-results})

\end{itemize}
\section{Implementation Details}

In this section, we provide the implementation details of our \ours.
% We didn't mention these details in our experiments because of the limited space. 
Our models are all built upon Detrection2\cite{wu2019detectron2}\footnote{\url{https://github.com/facebookresearch/detectron2}} and Detrex\cite{ren2023detrex}\footnote{\url{https://github.com/IDEA-Research/detrex}}. We keep most of the default hyper-parameters and use the standard data augmentation including random flip, random crop and scaling of the short edge in the range [400, 600] or [480, 800]. 
% scaling of the short edge in the range [480, 800] and set long edge 1333.
When we set the training iteration, with the number of datasets increasing, the total training iteration remains the same. As we mentioned on our main page, an exception arises with the ‘L+C+O+D’ dataset, where severe underfitting is observed when training only on O365~\cite{shao2019objects365} iterations. Consequently, we nearly double the training iterations.
\textcolor{black}{For fair comparison~\cite{unidet,scaledet,detectionhub}, we adopt ResNet50 pre-trained on ImageNet21K as the backbone when using Def-DETR~\cite{zhu2020deformable} detection architecture and pre-trained on ImageNet1K when using Sparse-RCNN~\cite{sparsercnn}. 
% For fair comparison~\cite{unidet,scaledet,detectionhub}, we adopt ResNet50 pre-trained on ImageNet21K as the backbone when comparing with Detic and SacleDet and pre-trained on ImageNet1K when using Sparse-RCNN~\cite{sparsercnn}
For training efficiency, in the ablation study, we set the batch size as 16 and train our models 180k iterations in the `C+L' setting and 450k iterations in the `C+L+O' setting.}
% To build the unified label space, we use label embedding generated by the pretrained clip model and modify it as our classifier for each dataset. 
% backbone? pretrained on ImageNet21k?
Other details are shown in Table~\ref{tab:implementation}. In Table~\ref{tab:datasets_url}, we provide the datasets we use in our experiments.
\label{appendix:impl_details}
\begin{table}[h]
    \centering
        \caption{\textbf{Implementation details.} Res50 denotes models are trained on ResNet50 backbone and * denotes the backbone is pretrained on ImageNet21k~\cite{russakovsky2015imagenet} otherwise is pretrained on ImageNet1k.}\label{table:increasing_multi}
        \tablestyle{2.5pt}{1}
        \begin{tabular}{lllccccc}\toprule
        Model & Datasets &Backbone & Iterations   & LR &Schedule &Batchsize \\\midrule
        \multirow{5}{*}{Baseline}
        & C &Res50* & 45k  &  1e-4&[40k,]& 64 \\
        & L &Res50* & 90k & 1e-4&[85k,] & 64 \\
        & O &Res50*& 450k &  1e-4&[350k,390k,]& 64 \\
        & D &Res50*& 450k&  1e-4&[350k,390k,]& 64 \\
        & C+L+O+D &Res50*& 800k & 1e-4&[740k,790k,]& 64\\\midrule
        
        \multirow{3}{*}{Def-DETR~\cite{zhu2020deformable}}
        & C+L &Res50*& 90k &  1e-4&[80k,]& 64 \\
        & C+L+O &Res50*& 450k &  1e-4&[350k,390k,]& 64\\
        & C+L+O+D &Res50*& 800k&  1e-4&[740k,790k,]& 64\\\midrule
         \multirow{1}{*}{Sparse-RCNN\cite{sparsercnn}}
         & C+O &Res50& 450k& 2.5e-5&[400k,440k,]& 16 \\
        % \multirow{1}{*}{Def-DETR~\cite{zhu2020deformable}}
        %  & C+L &Res50& 180k &  1e-4&[165k,]& 16 \\
        %  \multirow{1}{*}{(ablation)}
        %  & C+L+O &Res50& 400k&  1e-4&[350k,390k,]& 16 \\
        % \rowcolor[HTML]{efefef} \pretrain C+L+O+D & 49.2 & 37.4 &31.6 & 74.2 &\cellcolor[HTML]{efefef}-\\
        \bottomrule
        \end{tabular}
\label{tab:implementation}
\end{table}
\begin{table}[!ht]
   \centering
   \caption{\textbf{Dataset information. } The 4 datasets we use to train Plain-D$^{\text{N}}$et and the 13 datasets of ODinW. 
   % When counting the number of classes in downstream datasets, we exclude classes with an ID of 0. In ODinW13\cite{Li_2022_CVPR}, these classes correspond to the supercategories of the dataset, which are not annotated and are not utilized in our experiments.
   }
   \resizebox{\linewidth}{!}{%
   \begin{tabular}{lccc}
     \toprule
     Dataset & Classes&License &Website \\
     \midrule
      \multicolumn{3}{l}{\textit{\small{Training Datasets}}}\\\midrule
    COCO\cite{coco} &80&CC BY 4.0& \url{http://cocodataset.org} \\ 
     LVIS\cite{gupta2019lvis} &1203&CC BY 4.0& \url{https://www.lvisdataset.org} \\
     Objects365\cite{shao2019objects365}&365 &CC BY 4.0& \url{https://www.objects365.org} \\
     OpenImages\cite{oid} &601&CC BY 4.0& \url{https://storage.googleapis.com/openimages/web/index.html} \\\midrule
      \multicolumn{3}{l}{\textit{\small{Downstream Datasets}}}\\\midrule
     AerialMaritimeDrone& 5 & MIT & \url{https://public.roboflow.com/object-detection/aerial-maritime} \\
     Aquarium &7&CC BY 4.0& \url{https://public.roboflow.com/object-detection/aquarium} \\
     CottontailRabbits &1&CC BY 4.0&\url{https://public.roboflow.com/object-detection/cottontail-rabbits-video-dataset} \\
     EgoHands &1& CC BY 4.0&\url{https://public.roboflow.com/object-detection/hands} \\
     NorthAmericaMushrooms&2&Public Domain & \url{https://public.roboflow.com/object-detection/na-mushrooms} \\
     Packages&1& Public Domain &\url{https://public.roboflow.com/object-detection/packages-dataset} \\
     PascalVOC &20& CC BY 4.0 &\url{https://public.roboflow.com/object-detection/pascal-voc-2012} \\
     pistols&1 &Public Domain &\url{https://public.roboflow.com/object-detection/pistols} \\
     pothole&1&  ODbL v1.0 & \url{https://public.roboflow.com/object-detection/pothole} \\
     Raccoon &1& MIT &\url{https://public.roboflow.com/object-detection/raccoon} \\
     ShellfishOpenImages &3&CC BY 4.0 & \url{https://public.roboflow.com/object-detection/shellfish-openimages} \\
     thermalDogsAndPeople &2&Public Domain & \url{https://public.roboflow.com/object-detection/thermal-dogs-and-people} \\
     VehiclesOpenImages&5&CC BY 4.0 & \url{https://public.roboflow.com/object-detection/vehicles-openimages} \\
    % OpenImages & \url{https://storage.googleapis.com/openimages/web/index.html}
     \bottomrule
\label{tab:datasets_url}
   \end{tabular}
}
  %  \vskip -1em
  % \vskip -1em
\end{table}

\noindent\textbf{Training Datasets.}
To demonstrate the scalability and generalization of our method, a total of 17 object detection datasets were employed in our experiment. Detailed dataset descriptions are provided in the Supplementary. Here, we only provide a brief overview of the size and quantity of each dataset. (1) COCO~\cite{coco} has 80 object categories with nearly 0.1M images. (2) Object365v2~\cite{shao2019objects365} consist of 365 object categories. However, the latter includes approximately 1 million more images compared to the former. (3) LVIS~\cite{gupta2019lvis} has 1203 object categories with long tail distribution.  (4) OpenImages detection~\cite{oid} v6 has 601 object categories. (5) Object Detection in the Wild~\cite{zhang2022glipv2} 

\section{Detailed Results on ODinW}
\label{sec:detailed-ODinW}
As shown in Table~\ref{tab:odinw13}, we provide the fine-tuning transfer results on 13 downstream datasets. Due to the absence of official partitioning of training and testing sets in the Pistol dataset, we omit Pistol when computing the mAP across the downstream datasets. We achieve the best performance on swin-tiny\cite{liu2021swin} and resnet-50\cite{resnet} backbone. Compared with GLIP-T\cite{Li_2022_CVPR}, GLIPv2\cite{zhang2022glipv2}, Detic\cite{zhou2022detecting} and ScaleDet\cite{scaledet}, we obtain improvements of +2.5\%, +1.0\%, +8.8\%, and +1.0\%, in mAP across the 12 datasets. This experiment demonstrates that our model achieves optimal performance in fine-tuning on downstream datasets with great generalization ability.
\input{tables/table6(13)}
\section{Analysis on Proposal Generation}
 \label{sec:proposal-gen}
\begin{table*}[h]
  \centering
  \caption{{\bf The impact of using different label spaces from various datasets on generalization capability.}
  }
  % \vspace{-3mm}
  \resizebox{0.65\linewidth}{!}{%
  \begin{tabular}{c | ccc }
    \toprule
    Datasets& COCO~\cite{coco} & LVIS~\cite{gupta2019lvis} & O365~\cite{shao2019objects365}   \\
    \midrule
     Raccoon   & 30.4 & 56.3 & 58.1 \\
     thermalDogsAndPeople  & 0.2 & 55.3 &54.1 \\
     pistols   & 9.3 & 47.3 & 45.1  \\
     EgoHands  & 2.7 & 40.0 & 43.3   \\
    Aquarium   & 8.2 & 27.9 & 21.3 \\
    \midrule
    Mean & 10.2 & 45.4 & 44.4 \\
    \bottomrule
  \end{tabular}
  }
  % \vskip 1em
  
  \label{tab:proposal}
  % \vspace{-5mm}
\end{table*}

In this section, we provide an analysis of the impact of labels from different datasets on query generation. Table~\ref{tab:proposal} presents the impact of using different label spaces from various datasets on generalization capability (\eg, COCO denotes using label embeddings from COCO~\cite{coco} categories to generate query). As shown in Table~\ref{tab:proposal}, the best performance in downstream zero-shot settings is achieved with the label spaces from LVIS~\cite{gupta2019lvis} and O365~\cite{shao2019objects365} (45.3\% vs 44.4\%), which have relatively more categories in their label spaces. This demonstrates that a more general and unified semantic space can enhance generalization capability.

\section{Qualitative Results}

\newcommand{\fig}[2][1]{\includegraphics[width=#1\linewidth]{#2}}
\newcommand{\figh}[2][1]{\includegraphics[height=#1\linewidth]{fig/#2}}
\newcommand{\figa}[2][1]{\includegraphics[width=#1]{fig/#2}}
\newcommand{\figah}[2][1]{\includegraphics[height=#1]{fig/#2}}

%------------------------------------------------------------------------------

\begin{figure*}[!t]
% \vspace{-15mm}
\captionof{figure}{\textbf{Results for prediction on the COCO and LVIS.} Although COCO and LVIS share the same images, our \ours\text{} can predict different results with one weight thanks to flexible query generation. (Zoom in for a better view)
}
\centering
% \scriptsize
% \begin{tabular}{ccc}
% \fig[.4]{fig/fig1V5_p1} &
% \fig[.32]{fig/fig1V5_p2.pdf}&
% \fig[.2405]{fig/fig1V5_p3.pdf}
\fig[.99]{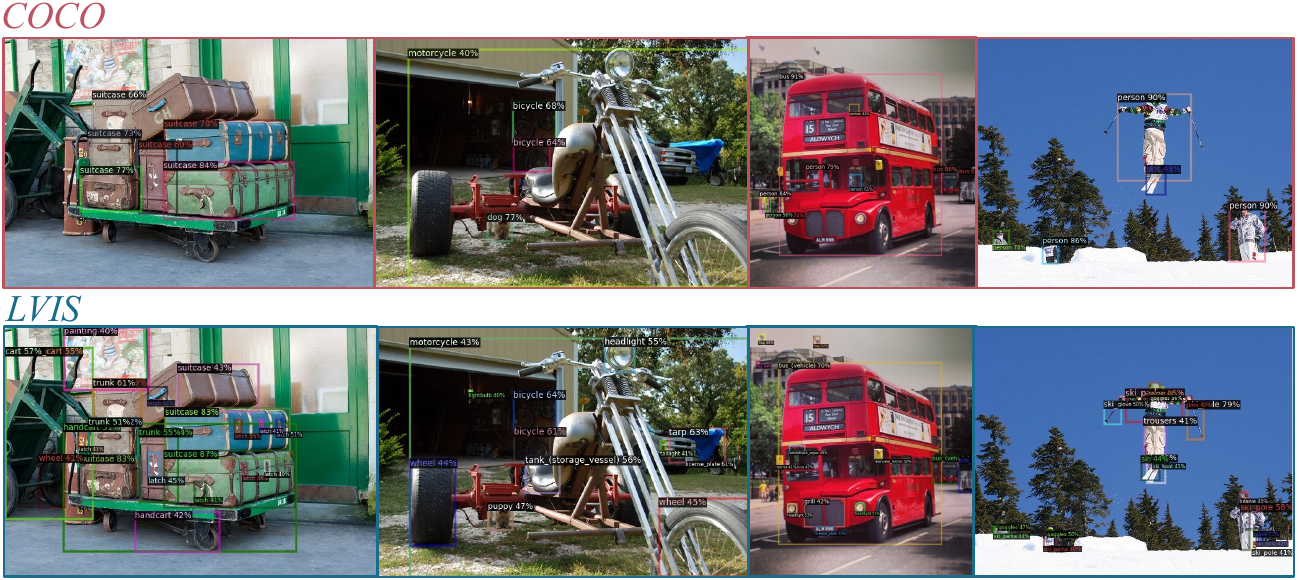}\\

% (a) & (b) & (c) \\
% \makecell{(a) Sampled Images and Features\\ in  \textcolor{coco}{COCO} \& \textcolor{obj}{Obj365}} &
% \makecell{(c) The performance-data volume \\ trade-off curve.}\\

% \end{tabular}%
% \vspace{-10mm}
\label{fig:figB}
% \par\vspace{24pt}
\end{figure*}
\begin{figure*}[h]
\captionof{figure}{\textbf{Qualitative results.} Results for prediction from \ours\text{}  on different domains and four datasets' validation sets.
}
% \vspace{-10mm}
% \scriptsize
% \begin{tabular}{ccc}
% \fig[.4]{fig/fig1V5_p1} &
% \fig[.32]{fig/fig1V5_p2.pdf}&
% \fig[.2405]{fig/fig1V5_p3.pdf}
\fig[.99]{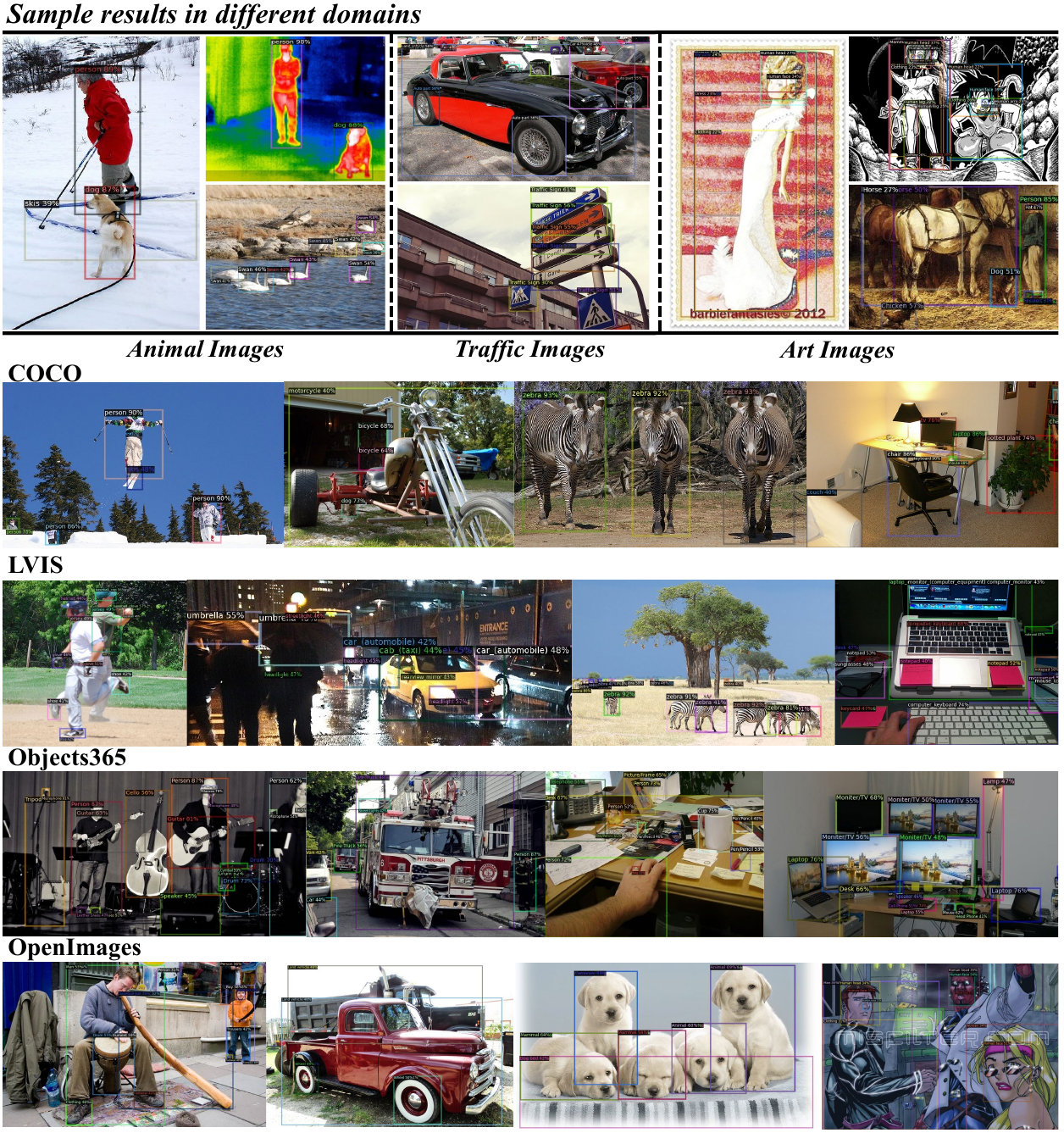}\\

% (a) & (b) & (c) \\
% \makecell{(a) Sampled Images and Features\\ in  \textcolor{coco}{COCO} \& \textcolor{obj}{Obj365}} &
% \makecell{(c) The performance-data volume \\ trade-off curve.}\\

% \end{tabular}%
% \vspace{-8mm}
\label{fig:figA}
% \par\vspace{24pt}
\end{figure*}

% \clearpage
\noindent
In this section, we provide the visualization results for prediction from our \ours. In Fig~\ref{fig:figB}, we use the same model weight to get the prediction results without further fine-tuning. Because the COCO and LVIS datasets share the same images, we demonstrate predictions across both datasets using identical images. It proves how our model effectively resolves conflicts arising from differing annotations on the same image across these two datasets. 
As shown in Fig~\ref{fig:figA}, we first demonstrate the robust detection capability of our model across different domains. Next, we showcase the visualization results on the validation sets of COCO, LVIS, Objects365, and OpenImages. 
% Corresponding to the seventh row in Table 2 of the main text, our model exhibits excellent performance on these datasets. 
\label{sec:qualitative-results}

% \newcommand{\fig}[2][1]{\includegraphics[width=#1\linewidth]{#2}}
% \newcommand{\figh}[2][1]{\includegraphics[height=#1\linewidth]{fig/#2}}
% \newcommand{\figa}[2][1]{\includegraphics[width=#1]{fig/#2}}
% \newcommand{\figah}[2][1]{\includegraphics[height=#1]{fig/#2}}

%------------------------------------------------------------------------------
% \vspace{-6pt}
% \clearpage

% \clearpage
\clearpage
\bibliographystyle{splncs04}
\bibliography{main}